\documentclass[journal]{IEEEtran}
\let\labelindent\relax
\usepackage{graphicx}
\usepackage{caption}
\usepackage{subcaption}
\usepackage{eqlist}
\usepackage{amsfonts}
\usepackage{tabularx}
\usepackage{booktabs}
\usepackage{amssymb}
\usepackage{amsmath}
\usepackage{enumitem}
\usepackage{threeparttable}
\usepackage[lined, ruled, linesnumbered, commentsnumbered]{algorithm2e}
\usepackage{lipsum}

\usepackage[usenames,dvipsnames]{xcolor}
\usepackage{flushend}

\newcommand{\emp}[1]{\textcolor{red}{#1}}

\begin{document}
%
\title{Planning Grasps for Assembly Tasks}

\author{Weiwei Wan, \IEEEmembership{Member, IEEE}, Kensuke Harada,
\IEEEmembership{Member, IEEE}, and Fumio Kanehiro, \IEEEmembership{Member, IEEE}
\thanks{Weiwei Wan and Kensuke Harada are with 
Graduate School of Engineering Science, Osaka University, Japan.
Fumio Kanehiro is with National Inst. of AIST, Japan.
E-mail: wan@sys.es.osaka-u.ac.jp}
}
\maketitle

\begin{abstract}
This paper develops model-based grasp planning algorithms for
assembly tasks. It focuses on industrial end-effectors like grippers and suction
cups, and plans grasp configurations considering CAD models of target objects.
The developed algorithms are able to stably plan a large number of high-quality
grasps, with high precision and little dependency on the quality of CAD models.
The undergoing core technique is superimposed segmentation, which pre-processes
a mesh model by peeling it into facets.
The algorithms use superimposed segments to locate contact points and parallel
facets, and synthesize grasp poses for popular industrial end-effectors.
Several tunable parameters were prepared to adapt the algorithms to meet various
requirements. The experimental section demonstrates the advantages of the
algorithms by analyzing the cost and stability of the algorithms, the precision
of the planned grasps, and the tunable parameters with both
simulations and real-world experiments. Also, some examples of robotic assembly
systems using the proposed algorithms are presented to demonstrate the efficacy.
\end{abstract}


\begin{IEEEkeywords}
Grasp synthesis, Grasp planning, Regrasp
\end{IEEEkeywords}

%
\IEEEpeerreviewmaketitle

\section{Introduction}
%
%
%
%
\IEEEPARstart{T}{his} paper develops algorithms to automatically plan
grasping poses for assembly tasks. It focuses on
industrial end-effectors, and plans grasp configurations for these
end-effectors considering CAD models of target objects.

Developing grasp planning algorithms is important to manufacturing
using ``teachingless'' robotic manipulators. Modern robotic manipulation systems use
manually annotated or pre-taught grasp configurations to perform certain tasks,
which is not only costly but also difficult to redeploy. Automatic grasp
synthesis or planning algorithms could bypass the annoying manual work and
enable fast redeployment for varying and changing manufacture. For this
reason, lots of studies in the field of robotic grasp have been devoted to automatic grasp planning
and many grasp planning algorithms have been developed. These algorithms 
are able to plan grasps considering forces and collisions. However, they hardly meet the
requirements of fully automatic manufacturing applications like bin-picking and assembly.
The requirements include but are not limited to:
(1) Large number of candidate grasps: The grasp planner is expected to provide a large number of
candidate grasp poses for optimization. (2) Stableness: The grasp planner must
have little dependency on the quality of CAD models. (3) Precision: Object poses
do not change much after being grasped by the planned grasps.

On the other hand, state-of-the-art grasp planning studies concentrate on
theoretical aspects like grasp closures and qualities, or applicational
aspects like grasping using
multi-finger hands, dexterous hands, and hands with tactile and force sensors.
Grasp planning for popular industrial end-effectors, e.g. parallel grippers and suction cups,
is usually ignored since
grasping rigid objects using suction cups and parallel grippers is considered to be easy
and solved. In this paper, we reinspect this opinion and
restudy the grasp planning problem for parallel
grippers. After reviewing previous work, we found that although
grasp planner for parallel grippers had existed for decades, they do not really
meet the requirements of industrial tasks: Some old-fashioned algorithms
could plan grasps for simple polytopes, but they cannot tackle complicated mesh models
and output satisfying number of grasps; Modern grasp
planning algorithms aim to find stable grasps. They cannot find a large amount of
grasping poses and don't consider about their precision. It is difficult to
use the planned grasps for industrial tasks like bin-picking and assembly, etc.

Under this background, this paper develops new grasp planning algorithms for
industrial bin picking and assembly. The problem setting is formulated as
follows.
The input includes:
(1) Kinematic models of industrial end-effectors like suction cups, parallel
grippers, and three-finger-one-parameter grippers; (2) Water-tight models of
rigid objects.
The output is:
A set of automatically planned grasp configurations. The algorithms assume:
(1) The objects have rigid bodies. Soft or changeable
objects are not considered; (2) The end-effectors and manipulators are actuated
using position control to ensure fast operation. Tactile or F/T sensors are not
available.

The undergoing core technique of the developed algorithms is superimposed
segmentation. The algorithms pre-process a mesh model by peeling it into
facets. Each facet is allowed to overlap with others and is thus called
superimposed segmentation. The overlap and sizes of segments are controlled by
several tunable parameters, which allow users to change the quality of planned
grasps following the requirements of their applications. The superimposed
segments are used to determine contact and compute parallel facets. Grasp
poses for popular industrial end-effectors are planned considering the contact
and parallel facets.

The developed algorithms could plan a large number of precise grasps with little
dependency on the quality of CAD models. Object positions change less than 2$mm$
after being grasped the planned grasps.
The computational cost and the stability of the algorithms in the presence of
low-quality CAD models, as well as the number of planned grasps and their
precision are analyzed in detail in the experimental section using both
simulational and real-world experiments.
The effect of the tunable parameters is analyzed by comparing the results of varying
values. A real dual-arm regrasp and assembly system is also implemented to show
the efficacy of the proposed algorithms.

The rest of the paper is organized as follows. Section II reviews related work.
Section III discusses the fundamental technique like superimposed segmentation and sampling
contact points. Section IV presents details of grasp planning algorithms
using the fundamental technique. Section V is the experimental section.
Section VI draws conclusions.

\section{Related Work}

This paper develops model-based grasp planning algorithms for suction cups and
parallel grippers. Accordingly, this section reviews related studies on
grasp theories and grasp planning of suction cups and parallel grippers, with
a special focus on the preprocessing of mesh models.

\subsection{Grasp theories and grasp planning}

Grasp theories study form/force closure and closure qualities. The theoretical
studies are applicable to suction cups, parallel grippers, as well as other
robotic hands.
Some early work like \cite{Salisbury82}\cite{Mishra87} studied point fingers and
polygonal objects, with later extensions to more realistic scenarios like curved
surfaces and fingers \cite{Nguyen86conf}\cite{Funahashi96}\cite{Rimon98a}\cite{Liu00}\cite{Stappen05},
considering grasp stability \cite{Montana91}\cite{Bicchi95}\cite{howard96} and grasp metrics
\cite{Mishra95}\cite{Roa15}. The early theoretical studies were mostly 2D, and
the concentration was to estimate the stability of grasps and the resistance to external wrenches.
The theoretical studies were extended to 3D polyhedral objects or mesh models composed of flat faces
later, assuming to be hard point contacts. Examples include \cite{Ponce97},
\cite{Liu04}, etc. Some other studies optimized the planned grasps
\cite{Fischer97} using some quality metrics \cite{Miller97}. 

Planning grasp poses for real-world objects and
real-world end-effectors are more challenging than the early theoretical studies. There is a big gap between
the computed results and real-world executions. One has to consider many factors
like contact regions, object surface curvatures, resistance to torque caused by
gravity forces, kinematics of robot hands, etc., to secure stable and
exact grasps. Several previous studies challenged these difficulties.
For example, Wolter et al. \cite{Wolter85} considered the geometry of grippers
during the automatic generation of grasps for 3D rectilinear objects. Jones et al.
\cite{Jones90} considered the parallel faces of a 3D object as well as the mesh
model of a robot gripper to plan two finger grasps for pick-and-place operations. Liu
et al. \cite{Liu14} used the attractive regions of an object to plan
stable grasps. Pozzi et al. \cite{Pozzi17} discussed grasp qualities
considering the kinematic structures of underactuated and compliant hands. Shi et al.
\cite{Shi17} considered about environmental constraints as well as the kinematic
constraints of robot hands to plan accessible grasps for bin-picking and kitting
tasks. Li et al. \cite{Li15} used stretching ropes (cord geometry) to find the
contact of a hand jaw with object surfaces and hence plan the grasps. Ciocarlie et al.
\cite{Ciocarlie07b} considered about local geometry and structures at contact
points and modeled friction forces using soft models. Harada et al.
\cite{Harada11} discussed about a gripper with soft finger pads attached to the
finger tips and analyzed object mesh models considering the depth of contacts.
These studies used gripper models and their kinematic structures to ensure
feasibility, and considered about contact properties by analyzing the meshes
around contact regions.

Our work plans grasp for suction cups and parallel grippers. For these simple
end-effectors, the form/force closure theory is
converted to comparing surface normals at the contacts. The kinematic
constraints, contact, and quality of grasps are considered in
segmentation, sampling, and nested collision detection. The quality of the
planned grasps depends on the preprocessing of mesh models, which is further
reviewed below.


\subsection{Preprocessing mesh models}

Two major approaches to preprocess the mesh models for grasp planning are (1)
primitive fitting and (2) surface segmentation. The first
approach represents mesh models using a set of shape primitives, and plans
grasp by considering the fitting errors or using pre-annotated grasps.
The second approach represents mesh models using coplanar
triangle sets.
Each coplanar segmented triangle set is named a facet and equals to one
constitutional polygon of a polyhedron. 

For primitive fitting-based grasp planning, Goldfeder et al.
\cite{Goldfeder07} represented a mesh model using recursive splitting and fitting of primitive
superquadrics \cite{Zha98}. El-Khoury et al. \cite{Ek07} fitted segmented point
clouds using primitive superquadrics and used pre-annotated training sets to learn grasp
points from the fitted models.
Xue et al. \cite{Xue09} also used primitive superquadrics to fit models and plan
grasps for Schunk Anthromorph Hands. 
Other than superquadrics-based fitting, Miller
et al. \cite{Miller03} represented a mesh model using a set of primitive mesh
models like boxes and spheres, and use a set of rules defined on the primitives
to generate grasps for the mesh model.
Hueber et al. \cite{Huebner08} fitted mesh models using different levels of
primitive boxes and planned grasp by evaluating the annotated grasps on the
primitives. Bonilla et al. \cite{Bonilla05} also fitted mesh models
using primitive boxes, and planned grasps using geometric information extracted from
the primitive boxes. Nagata et al. \cite{Nagata10} proposed an interactive
method for grasp planning by assuming shape primitives. Yamanobe et
al. \cite{Yamanobe10} defined the gripping configurations of several shape
primitives and used primitive shape representation to planning grasps for mobile
manipulators. Curtis et al. \cite{Curtis08} used primitives to learn grasps.
Instead of explicitly fitting primitives, the authors used learned grasping
knowledge on a set of primitive objects to speed up the process of planning
successful grasps for novel objects. Harada et al. \cite{Harada13} used
cylinders to fit banana point clouds, and planned robust grasps by analyzing the
projections of the point clouds on the cylindrical axes denoted by the fitted
cylinders. The grasp moduli space proposed by Porkorny et al.
\cite{Porkorny13} is also a primitive fitting approach. The primitive fitting
approaches do not assure the stability of planned grasps and exact object poses
after grasping. Some further evaluations or optimizations are needed to make the results
practical.

For surface segmentation-based grasp planning, Harada et al.
\cite{Harada11} clustered triangle meshes by using a parameter denoting
softness of contacts and implemented grasp planning for grippers with soft
finger pads. Tsuji et al. \cite{Tsuji14} used multi-level clustering
\cite{Garland01} to find the concavity and convexity of mesh models, and used
stress distribution models to plan stable grasps. Hang et al. \cite{Hang14icra}
also used multi-level clustering to plan grasps. The difference is their goal was not to
find grasping features. Instead, they use different levels of simplification for
iterative searching of stable grasps under reachability constraints. In a
later work, Hang et al. \cite{Hang14iros} extended the study to fingertip spaces
and used multi-resolution contacts to expedite grasp synthesis. The results
of the multi-level planning were demonstrated in \cite{Hang16} using an
Allegro hand. The hand could gait to different configurations as the weights of
objects change. Some of the primitive fitting approaches also have a
segmentation step, where meshes or point clouds are segmented for fitting
\cite{Ek07}\cite{Zhang03}.

The algorithms developed in this paper use surface segmentation to plan
contacts and estimate closure qualities. Unlike previous work which segmented
each triangle into a single facet or performed multi-level segmentation, the
algorithms allow superimposed segmentation. Each triangle is
repeatedly segmented into different facets, and the overlap and thickness of
facets are controlled by tunable parameters pertaining to surface normals. By
using sampled contact regions on the superimposed segmentations and leveraging
torque resistance, one may automatically plan a large number of stable and precise
grasp configurations for suction cups and parallel grippers.

\section{Pre-processing Mesh Models}

\subsection{Superimposed segmentation}

Superimposed segmentation provides uniform facets. Conventional approaches
\cite{Garland01}\cite{Kalvin96} cluster each triangle into a single facet,
resulting in uneven facets -- Some of them could be very large, while others are
very small. Unlike the conventional approaches, superimposed segmentation
allows one triangle to be repeatedly clustered to multiple facets, making all
facets uniform (equally large). The clustered facets do not exclusively
occupy the triangles that might also
belong to other facets.

\begin{figure}[!htbp]
  \centering
  \includegraphics[width = .97\linewidth]{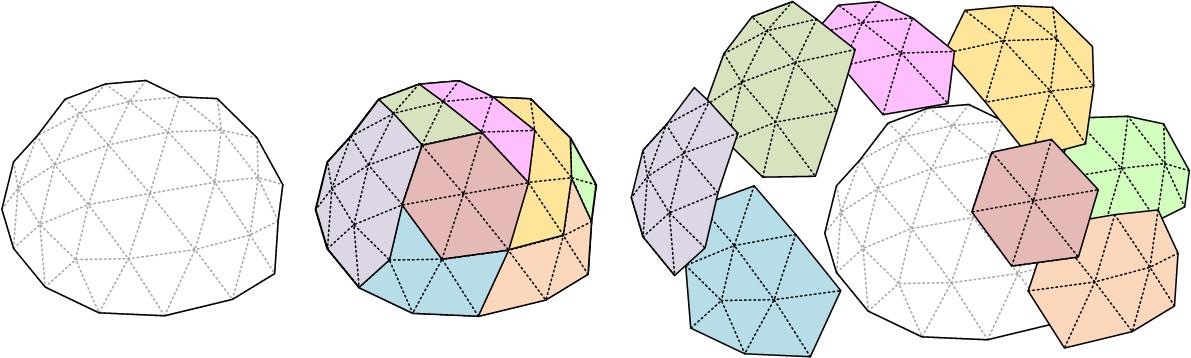}
  \caption{Segmenting mesh models into superimposed facets. Left:
  Original mesh model. Middle: Results of segmentation. Right: The facets are
  superimposed.}
\label{supersegs}
\end{figure}

\begin{figure}[!htbp]
  \centering
  \includegraphics[width = \linewidth]{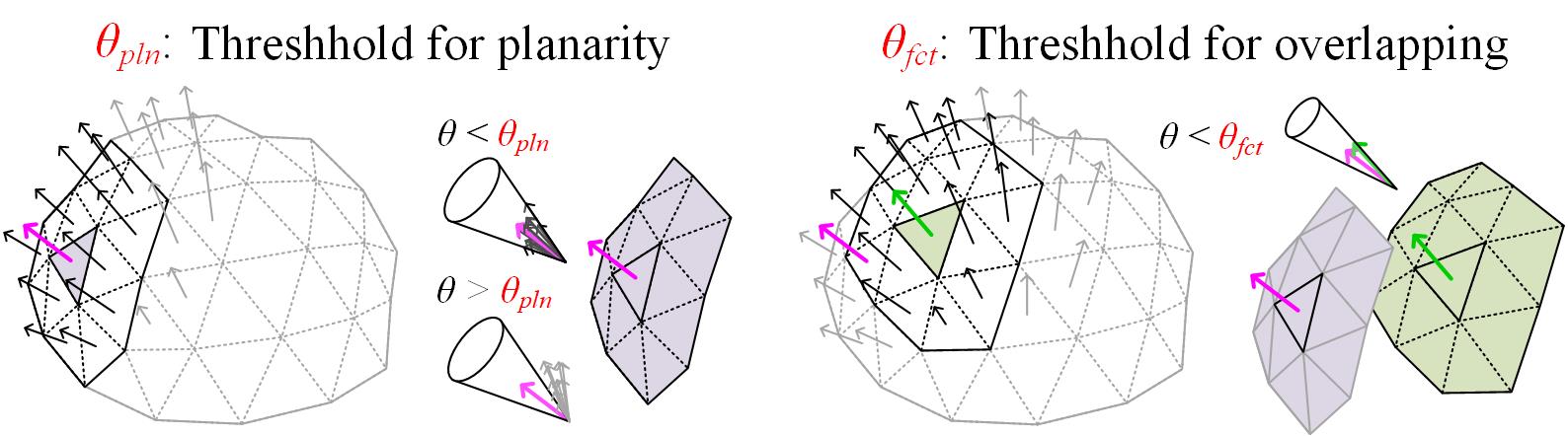}
  \caption{The superimposition is controlled by two parameters. (1)
  \emp{$\theta_{pln}$} controls planarity of each facet. (2)
  \emp{$\theta_{fct}$} controls the overlap of facets.}
\label{superparams}
\end{figure}

The superimposed segmentation is computed as follows. First, the algorithm
initiates a seed triangle and scans the surrounding triangles of the seed. See
the left part of Fig.\ref{superparams} for example. The purple triangle is the seed triangle,
and the algorithm scans the triangles surrounding it. If the angle
between the normal of the seed triangle and the normal of a nearby triangle is
smaller than a threshold \emp{$\theta_{pln}$}, the adjacent triangle is clustered into the
same facet as the seed triangle. In the figure, the angles between the black arrows
and the purple arrow are equal to or smaller than \emp{$\theta_{pln}$}, the related
triangles are clustered into the purple facet. In contrast, the angles between
the grey arrows and the purple arrow are larger than \emp{$\theta_{pln}$}, the related
triangles are not included. \emp{$\theta_{pln}$} is a
tunable parameter which controls the planarity of a facet.

After clustering the first facet, the algorithm initiates a new seed triangle and
repeats the clustering by starting from the new seed. The routine to initiate a new seed is as follows.
The algorithm scans the surrounding triangles of the previous seeds and checks the angles between the
normals of the previous seeds and the normals of the surrounding triangles. If an angle is
larger than \emp{$\theta_{fct}$}, the related triangle is selected as the new
seed. The right part of Fig.\ref{superparams} shows an example. The angle between the
green normal and the purple normal is larger than \emp{$\theta_{fct}$}. Thus, the green
the triangle with the green normal is selected as the new seed. The
algorithm repeats the clustering process by using new seed
and generates a new facet (the green facet shown in the right part
of Fig.\ref{superparams}). \emp{$\theta_{fct}$} is a tunable
parameter which controls the superimposition of facets.

During clustering, all triangles are repeatedly scanned, which allows one
triangle to be clustered into multiple facets. The facets could superimpose
with each other. On the other hand, the normals of all previous seeds are
compared when initiating a new seed, which ensures each facet to be unique
and the algorithm to stop properly.

\subsection{Sampling contact points}

\subsubsection{Sampling and distributing}
Contact points are computed by sampling the surface of the object mesh model.
The sampling is performed over the whole surface to provide evenly distributed contact points on the
mesh. After sampling, the sampled points are repeatedly distributed to the superimposed facets
as their contact points. Note that
we avoided sampling individual facets since it only provides
evenly distributed contact points on individual facets, the overall distribution
relies on segmenting methods. In cases where facets are small, individual
sampling may fail to produce contact points.

\begin{figure}[!htbp]
  \centering
  \includegraphics[width = \linewidth]{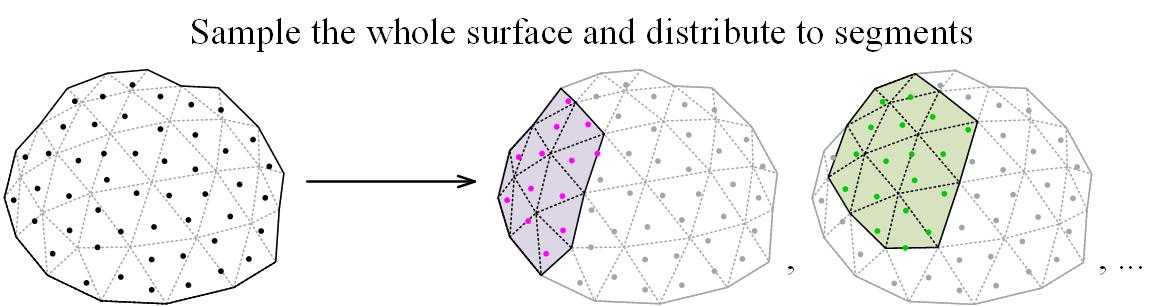}
  \caption{Left: Sample over the whole mesh surface. Right: Distribute the samples to
  each facet to avoid repeated sampling.}
  \label{samplingdistribute}
\end{figure}

Take Fig.\ref{samplingdistribute} for example. 
In the left part, the whole mesh surface is sampled. In the right part, the sampled points are
distributed to superimposed facets. The surface is sampled once and the sampled points
are distributed to individual facets repeatedly. The method ensures the contact 
points on each facet have equal density and are evenly distributed. It is irrelevant 
to the segmentation methods. Also, the method distributes the samples to
multiple facets without sampling again. It avoids repeated computation and improves algorithm efficiency.


\subsubsection{Removing bad samples}
The output of sampling and distributing cannot be used directly, since the distributed samples
might be (1) near the boundary of facets and (2) near to each other. In the
first case, attaching fingerpads to the sampled points near facet boundaries
lead to unstable grasps. In the second case,  attaching fingerpads to the near
points produces similar grasping configurations, which results in a large number of similar grasps
and is wasteful. To avoid the problems, we
perform two refining processes where the first one computes the distance between
a contact point and the boundary of its facet. The points with distances smaller
than \emp{$t_{bdry}$} will be removed. The second one removes the contact points that
are too close to others. The remaining contact points after being refined by the first process 
are further screened using the Radius Nearest Neighbour
(RNN) algorithm to remove nearby points with a distance smaller than
\emp{$t_{rnn}$}. Like \emp{$\theta_{pln}$} and \emp{$\theta_{fct}$},
\emp{$t_{bdry}$} and \emp{$t_{rnn}$} are tunable parameters of the grasp
planner.

\begin{figure}[!htbp]
  \centering
  \includegraphics[width = \linewidth]{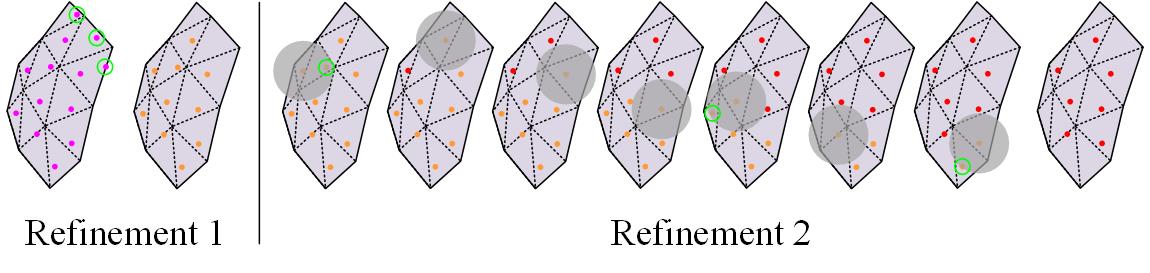}
  \caption{Two refinements that remove the bad contact points. The first refinement
  removes the contact points with small distances to facet boundaries. The second
  refinement removes the contact points that are too close to others.}
  \label{refinements}
\end{figure}

In practice, \emp{$t_{rnn}$} is determined by the size of finger pads. An
end-effector contacts with objects at a region, instead of a single point.
\emp{$t_{rnn}$} specifies the radius of the contact region. It controls the
density of planned grasps by removing nearby candidates.

\begin{figure}[!htbp]
  \centering
  \includegraphics[width = .9\linewidth]{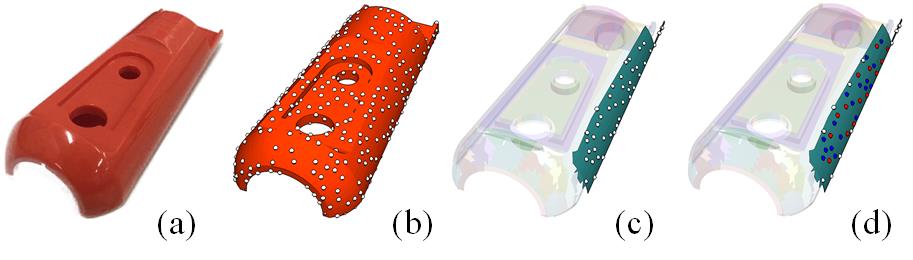}
  \caption{Sampling contact points. (a) The
  original object. (b) The sampled contact points. (c) The contact points distributed
  to one facet. (d) Removing bad contact points. Especially in (d), the white points
  are removed since they are too near to the boundary. The red points are the
  results of RNN screening.}
  \label{imposesample}
\end{figure}

As a demonstration, Fig.\ref{imposesample} shows the process of sampling contact points
using a plastic workpiece shown in
Fig.\ref{imposesample}(a).

\subsubsection{Stability}

The Soft-Finger Contact (SFC) model proposed in \cite{Harada14icra} is used to estimate the stability of
a grasp. The force and torque exerted by one SFC is expressed as:
\begin{equation}
\textbf{\textit{f}}_{t}^{T}\textbf{\textit{f}}_{t}+\frac{\tau_{n}^2}{e_{n}^2}\leq\mu^2f_{n}^2
\label{contactellipse}
\end{equation}

\begin{figure}[!htbp]
  \centering
  \includegraphics[width = \linewidth]{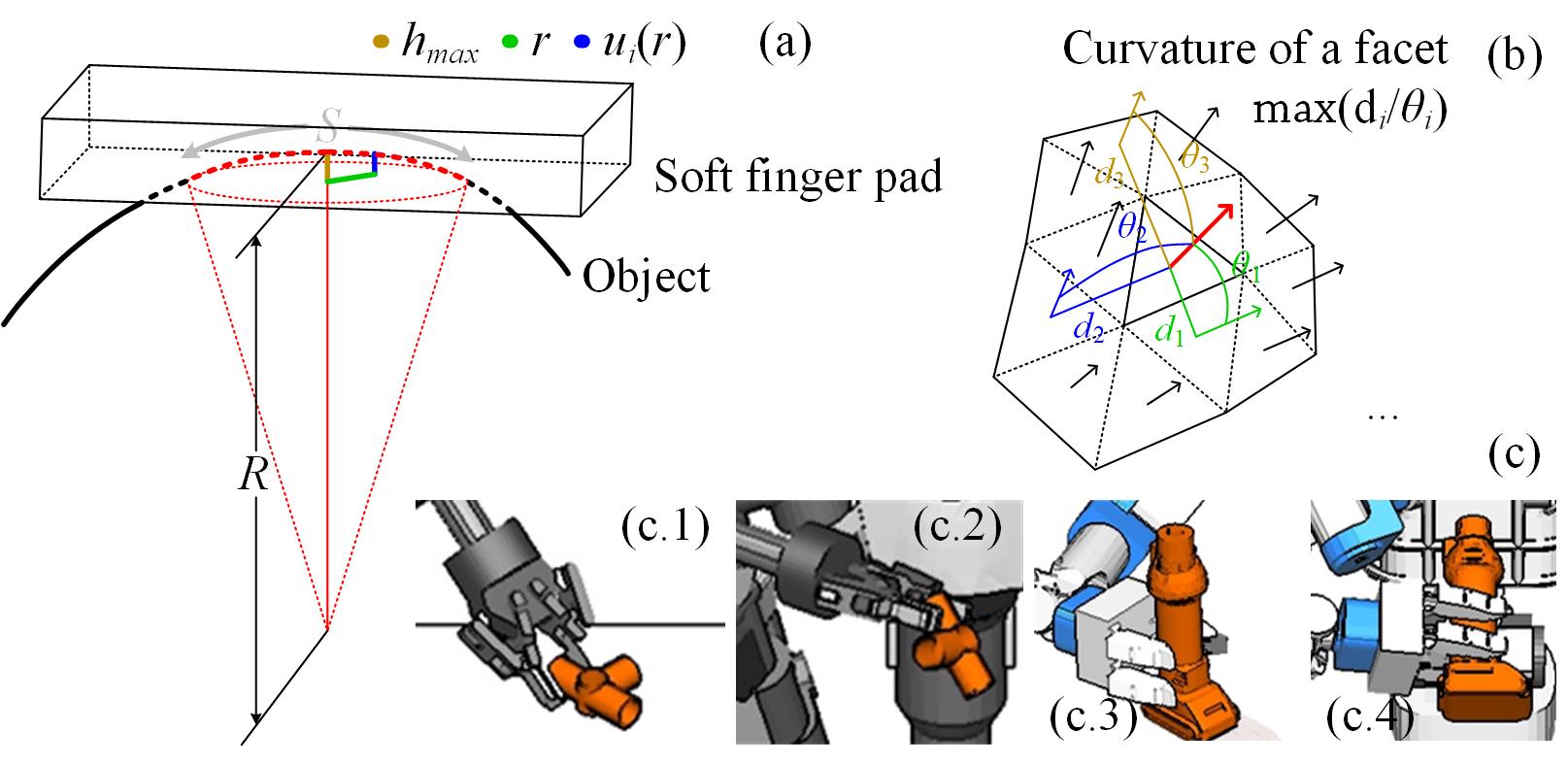}
  \caption{(a) The soft finger contact model. (b) Curvature of a facet.
  (c) The goal is to make sure the object is stable at an arbitrary pose.}
  \label{contact}
\end{figure}

Here, $\textbf{\textit{f}}_{t}$ indicates the tangential force at the contact.
$\tau_{n}^2$ indicates the torque at the contact. $f_{n}$ indicates the load 
applied in the direction of the contact normal. $e_{n}$ is the eccentricity parameter which
captures the relationship between maximum frictional force and moment. Under the
Winler elastic foundation model, $e_{n}$ is
\begin{equation}
e_{n}=\frac{\mathtt{max}(\tau_n)}{\mathtt{max}(f_t)}=\frac{\int_S r\mu K u_i(r)dS}{\int_S \mu K u_i(r)dS}
\label{eccentricity}
\end{equation}
where $K$ is the elastic modules of the foundation over the thickness of the soft finger pad.
$S$ is the contact surface between the finger pad and the object. 
$r$ is the distance between a differential contact point and the center of the contact region.
$u_i(r)$ is the depth of the soft penetration. These symbols are illustrated in Fig.\ref{contact}.

Using $R$ to denote the radius of the the contact curvature,
$u_i(r)$ can be represented by
\begin{equation}
u_i(r) = \sqrt{R^2-r^2}-(R-{\color{red}h_{max}})
\label{uir}
\end{equation}
where $h_{max}$ is the maximum depth of the soft penetration.
Following the definition of curvature, $R$ could be computed using
$\mathtt{max}(d_/\theta_i)$, where $d_i$ is the distance between the 
center of the $i$th in the facet and the center of the seed
triangle, $\theta_i$ is the angle between the normal of the $i$th triangle
and the normal of the seed triangle. The computation is illustrated in Fig.\ref{contact}(b). 
$R$ is essentially determined by the geometry of a facet.
${\color{red}h_{max}}$ is used as a tunable parameter to control the stability of planned grasps.
Since the goal of stability estimation is to make sure the object is stable at an arbitrary pose (Fig.\ref{contact}(c)),
a planned grasp must meet\footnote{Here, we are considering the worst case where the gripping
torque must resist the largest torque caused by gravity. During
manipulation, the largest external torque appears when
gravity direction is perpendicular to
vector $\overrightarrow{contact-com}$.}
\begin{equation}
(mgc)^2\leq{e_n}^2(\mu^2f_{n}^2-(mg)^2)
\label{stbcond}
\end{equation}
where $c$ is the distance between the $com$ (center of mass) of the object and the center of contacts.
From (\ref{eccentricity}), (\ref{uir}), and (\ref{stbcond}) we obtain
\begin{equation}
(mgc)^2\leq
(\frac{8}{15})^2(2R{\color{red}h_{max}}-{\color{red}h_{max}}^2)(\mu^2f_{n}^2-(mg)^2)
\end{equation}
This equation is used to determine the stability of planned grasps.

\section{Planning the Grasp Configurations}

Using the superimposed facets and the sampled contact points, we develop
algorithms to plan grasps for suction cups and parallel grippers.

\subsection{Suction cups}

\begin{figure}[!htbp]
  \centering
  \includegraphics[width = \linewidth]{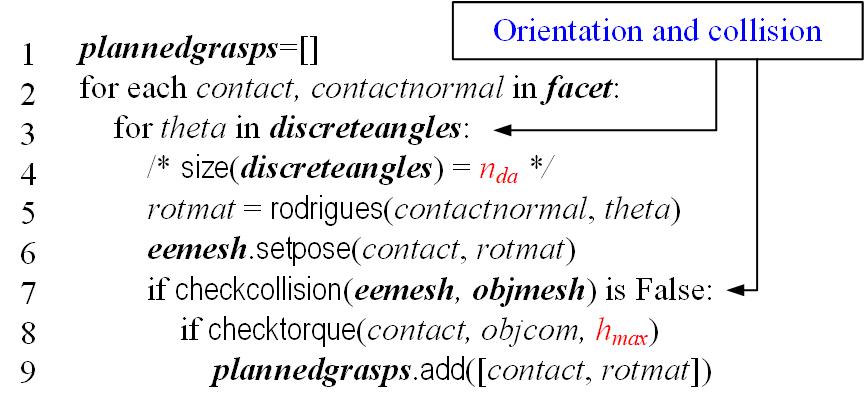}
  \caption{The pseudocode of grasp planning for suction cups. It is a two step process where
  the first step finds possible orientations and performs $\mathtt{checkcollision}$, the second step
  performs $\mathtt{checktorque}$.}
  \label{suctioncup}
\end{figure}

We assume a suction end-effector has only one suction cup. The grasp planning
algorithm for a single suction cup is a two-step process. In the first step, the
algorithm finds the possible orientations to attach the suction cup to the
sampled and refined contact points. Since the approaching direction must be
perpendicular to the contact region, the end-effector's orientation is only
changeable by rotating around the approaching direction. The algorithm poses the
suction cup to the contact points from the changeable orientations and removes
the infeasible (collided) grasps. In the second step, the algorithm
examines the resistance of planned suction configurations to external torques
caused by gravity. The pseudocode of the algorithm is shown in
Fig.\ref{suctioncup}.

For each contact point, the algorithm discretizes the rotation around the
contact normal into $\textbf{\textit{discreteangles}}$, and computes the
rotation matrices. The number of discretized values is determined by
\emp{$n_{da}$}. The algorithm poses the $\textbf{\textit{eemesh}}$ (mesh model of the end effector)
using the computed rotation matrices and checks the collision between the
$\textbf{\textit{eemesh}}$ and the $\textbf{\textit{objmesh}}$ (mesh model of
the object).  Line 3 of the pseudocode iterates through the discrete
orientations. Line 6 poses the $\textbf{\textit{eemesh}}$ to a $contact$ with
rotation matrix $rotmat$. Line 7 checks the collision between
$\textbf{\textit{eemesh}}$ and $\textbf{\textit{objmesh}}$.

In the second step, the algorithm further examines the resistance to external
torques. The function $\mathtt{checktorque}$($contact$, $objcom$, ${\color{red}h_{max}}$) in line 8 of
the pseudocode performs the examination. The function computes the Euclidean distance between $contact$ and
$objcom$ (the $com$ of the object), and checks if Eq.\eqref{stbcond} is met.

\begin{figure}[!htbp]
  \centering
  \includegraphics[width = 0.85\linewidth]{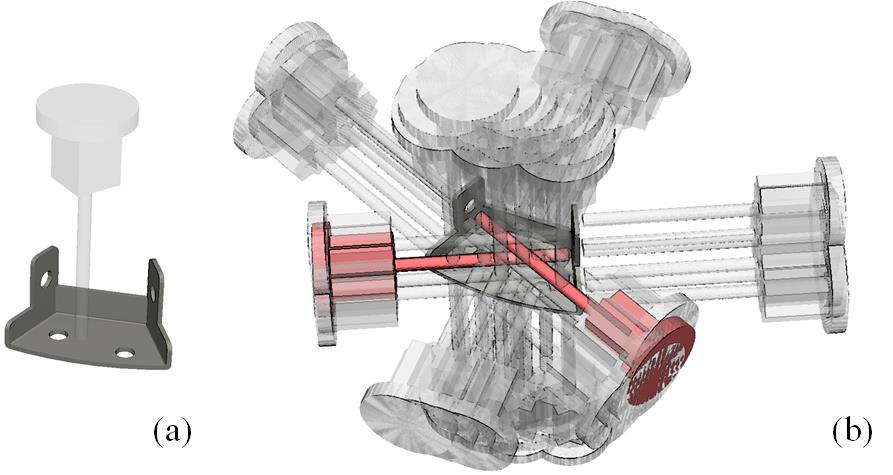}
  \caption{Results of grasp planning for a suction cup
and a metal workpiece. (a) One of the planned grasps. (b) All results.
The red grasps collide with the metal workpiece.}
  \label{suctioncupdemo}
\end{figure}

Fig.\ref{suctioncupdemo} shows the results of grasp planning for a suction cup
and a metal workpiece. A single grasp is shown in
Fig.\ref{suctioncupdemo}(a). All planned results are shown in
Fig.\ref{suctioncupdemo}(b). The red configurations in
Fig.\ref{suctioncupdemo}(b) are the grasps deleted by the
$\mathtt{checkcollision}$($\textbf{\textit{eemesh}}$, $\textbf{\textit{objmesh}}$)
function.

\subsection{Two-finger parallel grippers}

Grasp planning for two-finger parallel grippers is a three-step process. In the
first step, the planner for two-finger parallel grippers first finds parallel
facets and computes candidate contact pairs by examining the contact points on
the parallel facets. Compared with suction cups which need one contact point,
the planner for two-finger parallel grippers needs two contact points with
opposite contact normals. Thus, the algorithm involves an extra step to prepare
the candidate contact pairs. The second and third steps are similar to suction
cups. In step two, the algorithm finds the possible orientations to attach the
parallel gripper to the candidate contact pairs. In step three, the algorithm
examines the stability of the planned grasps.

\begin{figure}[!htbp]
  \centering
  \includegraphics[width = \linewidth]{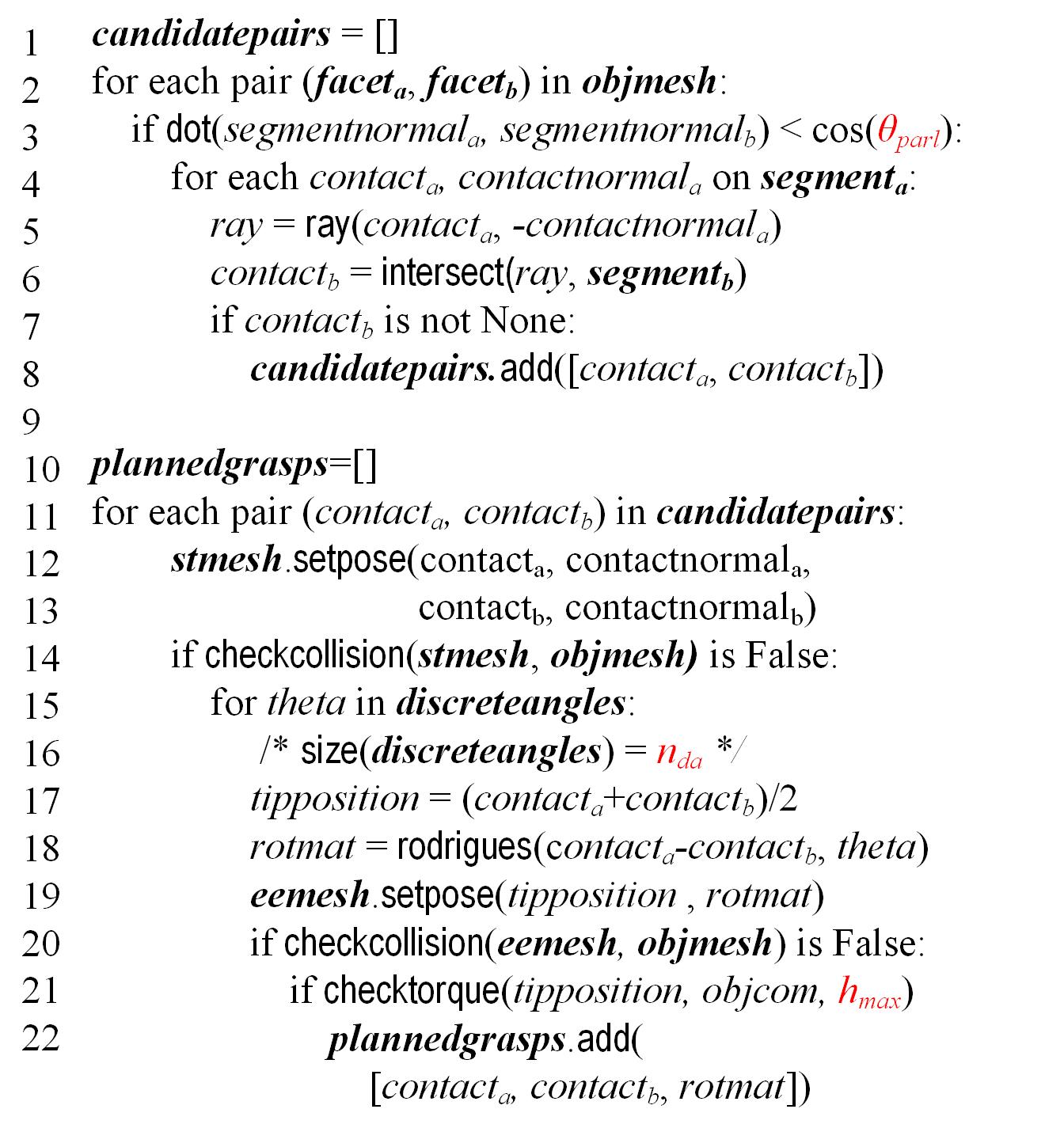}
  \caption{Pseudocode of grasp planning for two-finger parallel grippers.
  \textbf{\textit{stmesh}} denotes the collision model of a gripper's stroke.
  \textbf{\textit{eemesh}} denotes the collision model of a gripper.}
  \label{twofingergripper}
\end{figure}

The pseudocode of the grasp planner for two-finger parallel grippers is shown in
Fig.\ref{twofingergripper}. In the first block (lines 1 to 8), the algorithm
finds candidate contact pairs. For each pair of parallel facets, the algorithm
initiates a ray that starts from a contact point on one facet and points to the
inverse direction of the contact normal. It detects the intersection between the ray and
the other facet. If an intersection exists, the contact-intersection pair is saved as a candidate.
Whether two facets in a pair are parallel is determined by a tunable parameter
\emp{$\theta_{parl}$} (see line 3). In the second block (lines 10 to 22), the
algorithm performs collision detection and examines stability. The algorithm
invokes two nested $\mathtt{checkcollision}$ functions in the second block. In
the first call (line 14), the algorithm checks if the stroke of the gripper
collides with the object. Stroke is represented by cylinders which do not have orientation. 
In the second call (line 20), the algorithm checks if the whole hand (both fingers and
palm) collides with the object. The algorithm poses the fingertips to the
center of the contact pairs and performs collision at different rotations
around the axis formed by the two contacts. During the process, the planner
attaches one finger pad to one position of a contact pair, and attaches the other finger
pad to the other position of the contact pair. Suppose the two parallel fingers
are $f_1$, $f_2$, the contact pair is [$contact_a$, $contact_b$], the algorithm
attaches $f_1$ to $contact_a$, attaches $f_2$ to $contact_b$, rotates the gripper around the axis
$\overrightarrow{contact_a-contact_b}$, and checks the collision between
$\textbf{\textit{eemesh}}$ and $\textbf{\textit{objmesh}}$ at every
orientation (lines 15 to 20).

\begin{figure}[!htbp]
  \centering
  \includegraphics[width = .85\linewidth]{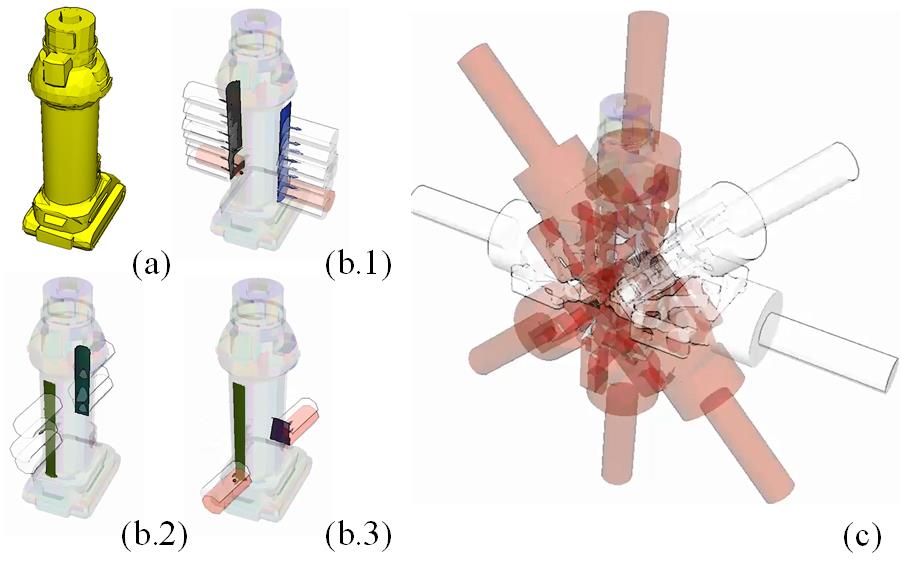}
  \caption{Results of grasp planning for a Robotiq85 gripper
and an electric drill. (a) The CAD model of the drill. (b.1-3)
Some of the parallel facets and the results of
$\mathtt{checkcollision}(\textbf{\textit{stmesh}},\textbf{\textit{objmesh}})$.
The cylinders indicate the stroke of the gripper. The red
ones collide with the drill. The white ones are collision-free.
(b) The discretized grasps at one contact pair. The red hands indicate the
obstructed grasps. The white ones
are the planned grasps.}
  \label{graspsdemo}
\end{figure}

Fig.\ref{graspsdemo} shows the results of grasp planning for a Robotiq85 gripper
and an electric drill. The CAD model of the drill is shown in
Fig.\ref{graspsdemo}(a). The collision between strokes and the model is
detected at line 14 of Fig.\ref{twofingergripper}. Fig.\ref{graspsdemo}(b.1-3)
draw some results of the collision detection.
The red cylinders show the strokes that collide with the object and
are removed to avoid repeated collision checking at different orientations. The
white cylinders are further examined in line 20 of Fig.\ref{twofingergripper} to see
if there is a collision between the whole hand and the object.
The discretized orientations at one contact pair and the
results of whole-hand collision detection are illustrated in Fig.\ref{graspsdemo}(c). The red
hands indicate the collided grasps found by
$\mathtt{checkcollision}(\textbf{\textit{eemesh}},\textbf{\textit{objmesh}})$. The white ones
are the planned grasps.

\subsection{Three-finger parallel grippers}

In addition, the planner proposed in this paper is applicable to three-finger
parallel grippers where two fingers are actuated together against a
third finger.

\begin{figure}[!htbp]
  \centering
  \includegraphics[width = \linewidth]{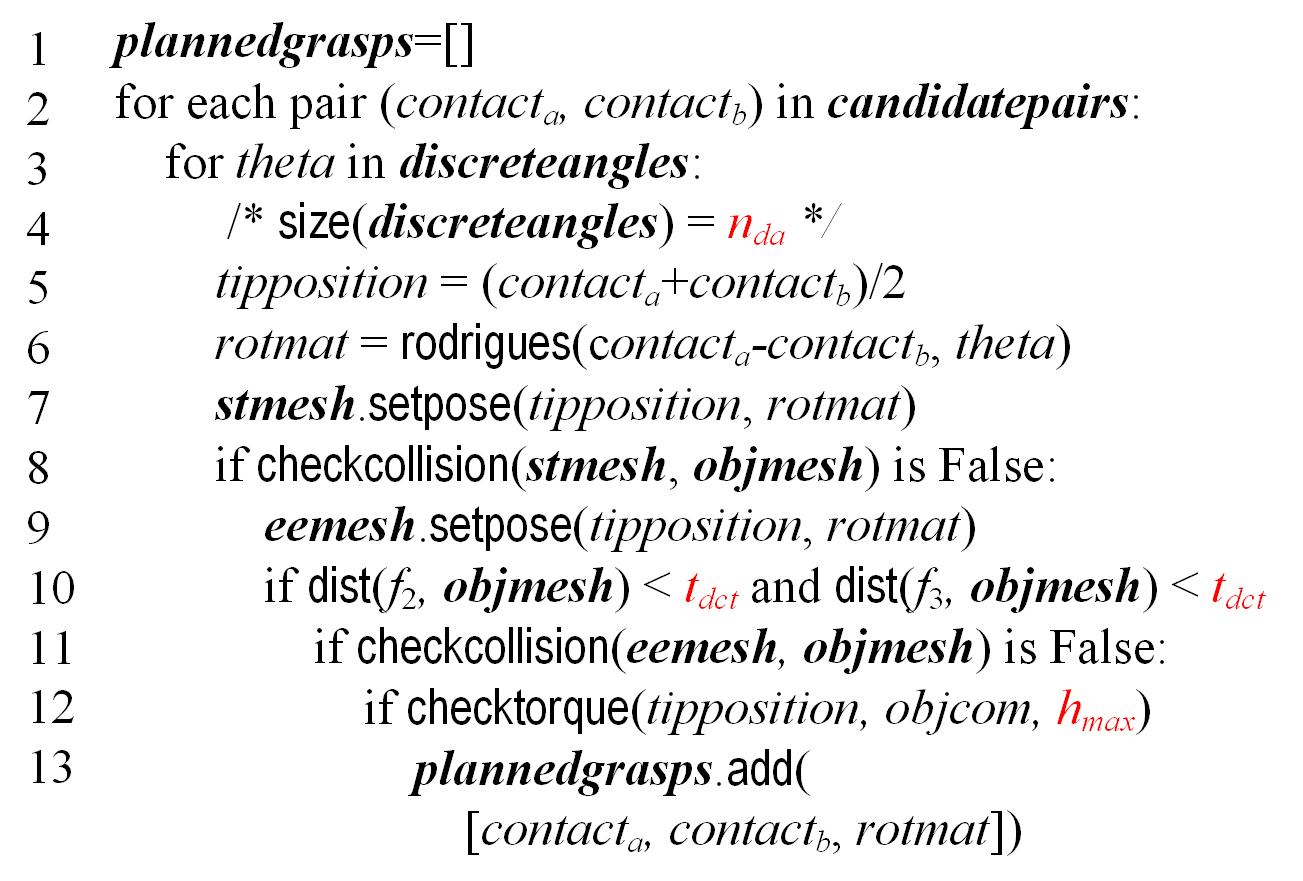}
  \caption{Pseudocode of grasp planning for three-finger parallel grippers.}
  \label{threefingergripper}
\end{figure}

\begin{figure}[!htbp]
  \centering
  \includegraphics[width = .85\linewidth]{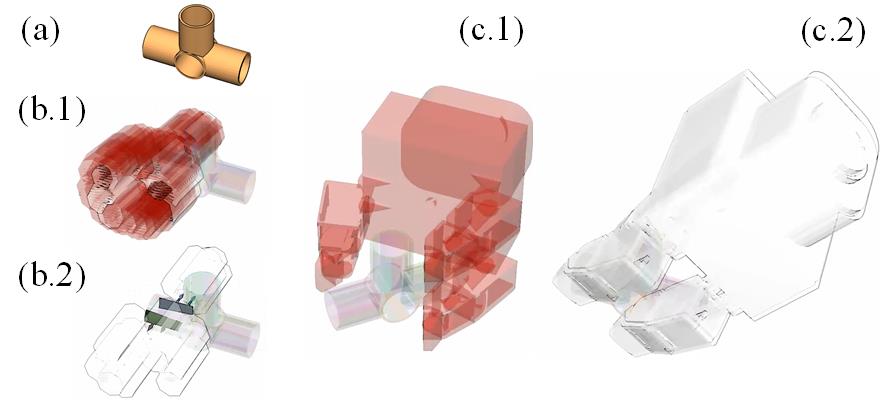}
  \caption{Results of grasp planning for a three-finger gripper
and a tube connector. (a) Mesh model of the object. (b.1) Results of the
collision detection on strokes. (b.2) A clear view of the collision-free strokes
of (b.1). (c.1) A collided grasp configuration found by the second collision
checking. (c.2) A collision-free grasp configuration (it is also one planned
grasp configuration).}
  \label{threegraspsdemo}
\end{figure}

The process is similar to grasp planning for two-finger parallel grippers,
except that the two fingers on one side of the parallel gripper are treated as a
single finger. Its pseudocode is shown in Fig.\ref{threefingergripper}.
In the first step, the planner finds parallel facets and computes candidate
contact pairs (not shown in Fig.\ref{threefingergripper}). In the second step,
the planer poses the fingertips to the center of the contact pairs and performs
collision detection at different rotations around the axis formed by the two contacts
(Fig.\ref{threefingergripper}).
The algorithm also invokes two nested $\mathtt{checkcollision}$ functions.
The first invocation checks the collision of strokes (line 8). The second
invocation checks the collision of the whole hand (line 11). However, different
from the two-finger case, the stroke of a three-finger gripper is represented by three
cylinders, which changes with hand orientation. Thus, both the two
invocations are performed under a specific orientation (both are invoked inside the initialized by
line 3). If the stroke is collision free,
the planner attaches one finger pad to one position of a contact pair and
attaches the center of the other two finger pads to the other position of the
contact pair (line 9). Suppose the three fingers are $f_1$, $f_2$, and $f_3$
where $f_1$ is at one side of a parallel gripper, $f_2$ and $f_3$ are on the other side. The contact
pair is named [$contact_a$, $contact_b$].
The algorithm attaches $f_1$ to $contact_a$, attaches
$\frac{f_2+f_3}{2}$ to $contact_b$, and examines the distances between $f_2$
and $\textbf{\textit{objmesh}}$, and $f_3$ and
$\textbf{\textit{objmesh}}$ (line 9). The algorithm requires
$||f_2$-$\textbf{\textit{objmesh}}||$$<$\emp{$t_{dct}$} and
$||f_3$-$\textbf{\textit{objmesh}}||$$<$\emp{$t_{dct}$} where \emp{$t_{dct}$} is a
tunable parameter.
The parameter controls the contact between the two fingers and the object
surface. If the distances meet the requirements, the algorithm checks the
collision between the gripper and the object (line 10). In the third step,
the algorithm examines the stability (line 12).

Fig.\ref{threegraspsdemo} shows the results of grasp planning for a three-finger
gripper and a tube connector. Some exemplary results of collision detection with
strokes are illustrated in Fig.\ref{threegraspsdemo}(b.1-b.2). Fig.\ref{threegraspsdemo}(b.1)
shows both the collided (red) and collision-free (white) strokes. The two-finger
side is in the front.
The strokes are rotated around the axis passing through the contact pair. 
Fig.\ref{threegraspsdemo}(b.2) is a clear view without the collided ones. The
collision-free strokes are further examined in the second collision detection. Some
exemplary results are illustrated in Fig.\ref{threegraspsdemo}(c.1) (a collided grasp
configuration) and Fig.\ref{threegraspsdemo}(c.2) (a collision-free grasp
configuration).

\section{Analysis and Demonstrations}

\subsection{Tunable parameters}

In the algorithms, seven tunable parameters are prepared for user configuration. 
The parameters and their functions are
shown in Table.\ref{parameters}.
This section analyzes the parameters and compares the
performance of different parameter settings by comparing the different planned results.
In practice, users may set the parameters according
to the needs of their robotic systems.
\begin{table*}[!htbp]
\centering
\caption{The tunable parameters}
\begin{threeparttable}
\begin{tabular}{l|l|l}
\toprule
 Name & Function & Where
 \\
 \midrule
 \midrule
 \emp{$\theta_{pln}$} & Control the planarity of each facet & Appeared in
 superimpose segmentation\\
 \emp{$\theta_{fct}$} & Control the overlap of facets & Appeared in
 superimpose segmentation\\
 \emp{$t_{bdry}$} & Control the distance between contacts and facet
 boundaries & Appeared in removing bad samples\\
 \emp{$t_{rnn}$} & Control the radius of contact regions & Appeared in removing
 bad samples\\
 \emp{$h_{max}$} & Control the stability of the planned grasps & Appeared in all grasp planners\\
 \emp{$\theta_{parl}$} & Control the parallelity of two facets & Appeared
 in the planners for two and three-finger grippers\\
 \emp{$t_{dct}$} & Control the distances between finger pads and object
 surfaces & Appeared in the planner for three-finger grippers\\
 \emp{$n_{da}$} & The number of discretized rotation angles around contact
 normals & Appeared in all grasp planners\\
 \midrule
\bottomrule
\end{tabular}
\end{threeparttable}
\label{parameters}
\end{table*}

\begin{figure}[!htbp]
  \centering
  \includegraphics[width = \linewidth]{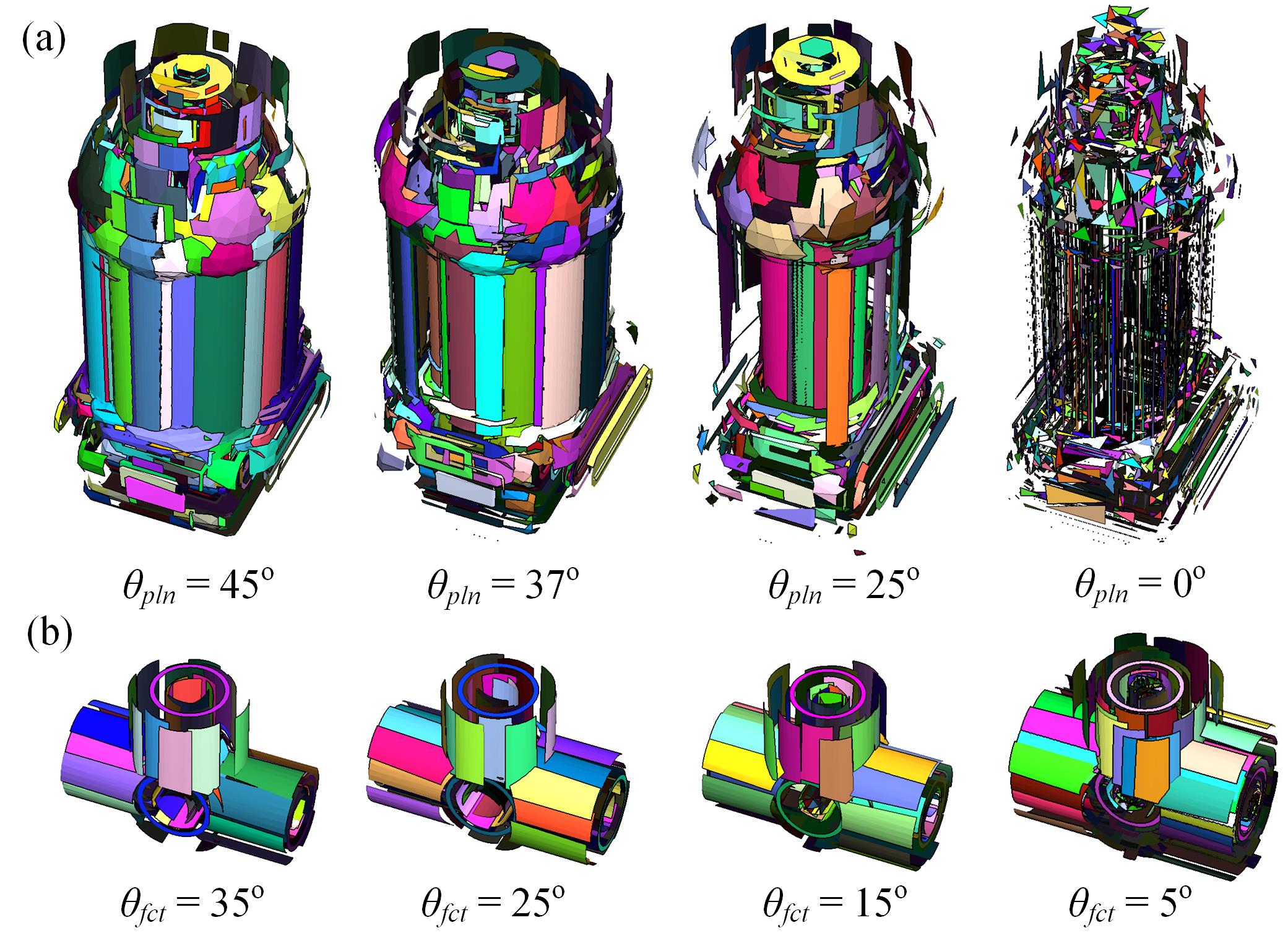}
  \caption{(a) Results of superimposed segmentation using different
  $\theta_{pln}$. Each segment is shown in a random
  color, and is drawn with a random offset from
  its original position to give a clear view. The results show that
  a smaller $\theta_{pln}$ leads to flatter and smaller facets. (b) Results of superimposed segmentation using different
  $\theta_{fct}$. The object used for demonstration is the one in
  Fig.\ref{threegraspsdemo}(a). The results show that a smaller $\theta_{fct}$ leads
  to more overlap.}
  \label{edsegresults}
\end{figure}
 
\begin{figure}[!htbp]
  \centering
  \includegraphics[width = .99\linewidth]{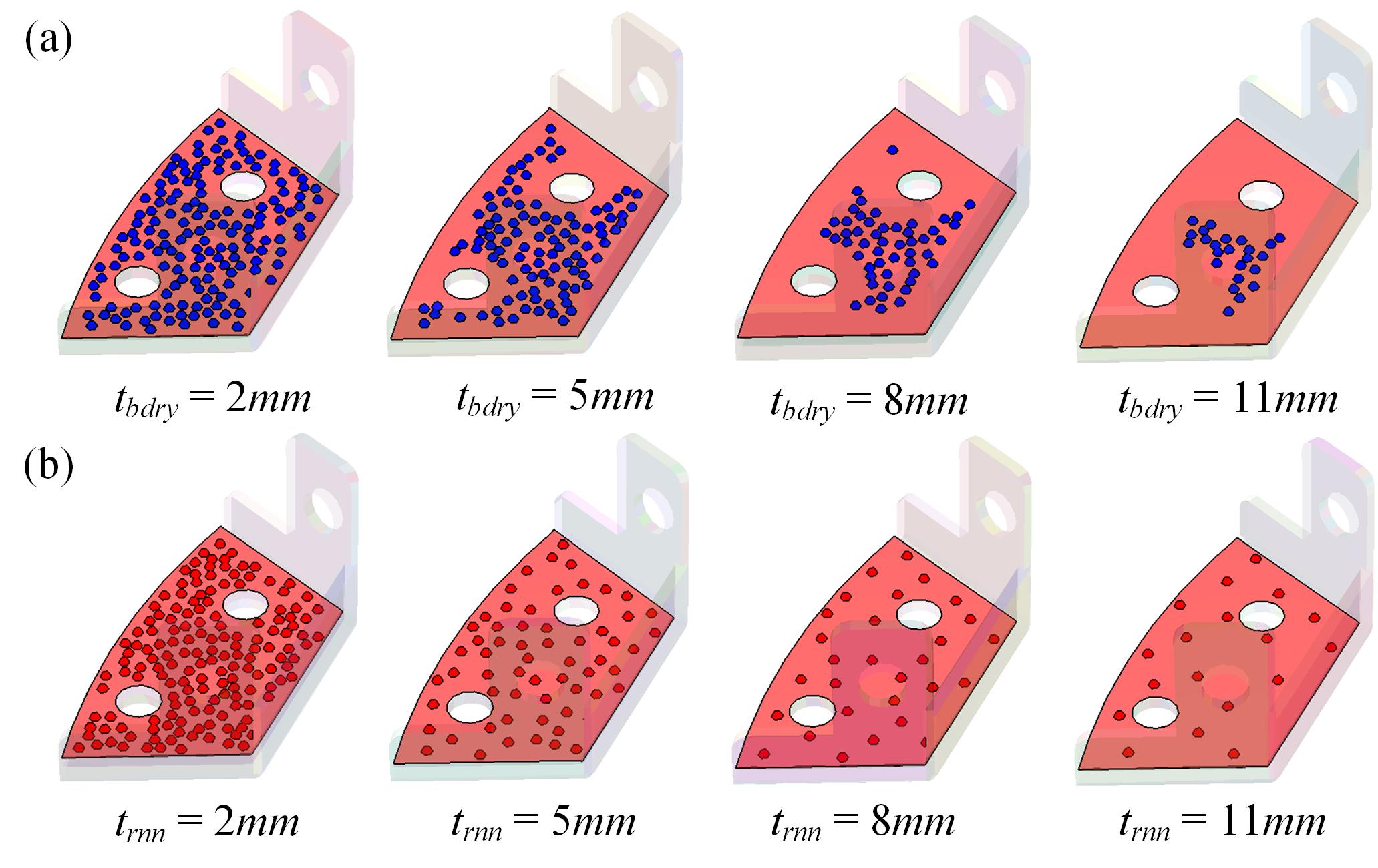}
  \caption{(a) Results of contact sampling using different $t_{bdry}$.
  The resulting samples are drawn in blue color.
  The results show that a smaller $t_{bdry}$ leads to a smaller clearance between
  contact samples and facet boundaries.
  (b) Results of contact sampling using different $t_{rnn}$.
  The resulting samples are drawn in red color. The results show that a smaller $t_{rnn}$
  leads to denser contact samples.}
  \label{edsegresults1}
\end{figure}

\begin{figure*}[!htbp]
  \centering
  \includegraphics[width = \linewidth]{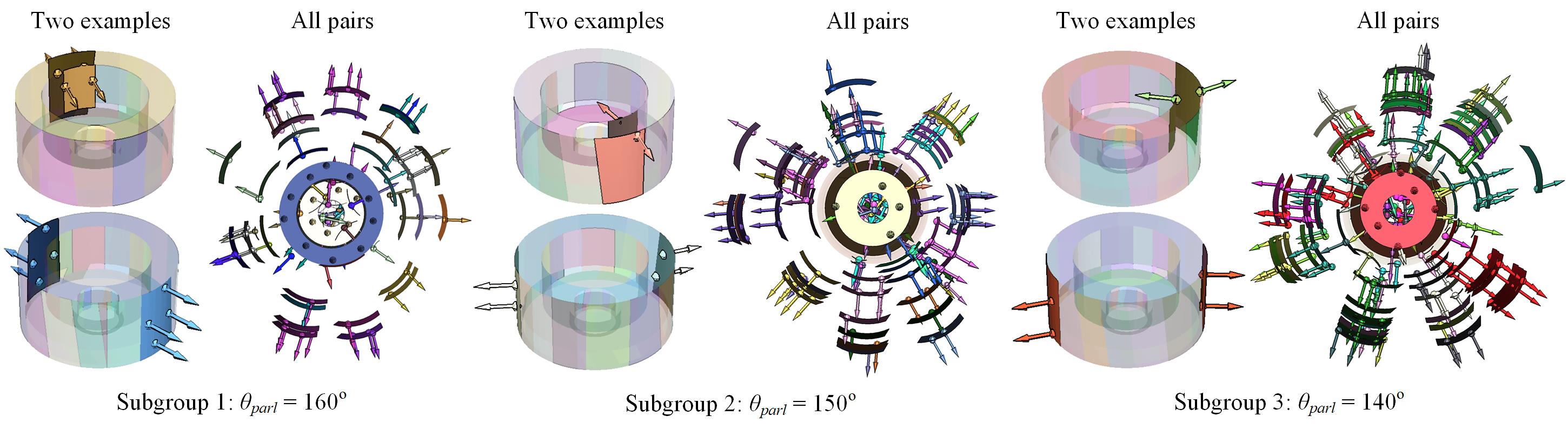}
  \caption{Results of parallel facets using
  different $\theta_{parl}$. Arrows indicate facet normals. Two facets with
  the same arrow color are parallel. The left part of each subgroup shows two
  exemplary pairs. The results show that as $\theta_{parl}$ decreases,
  less parallel facets are accepted.
  The right part of a subgroup shows all pairs. Here, the facets are drawn
  with random offsets from their original position to give a good view.
  As $\theta_{parl}$ decreases, more pairs are found.}
  \label{edsegresults2}
\end{figure*}

\begin{figure}[!htbp]
  \centering
  \includegraphics[width = \linewidth]{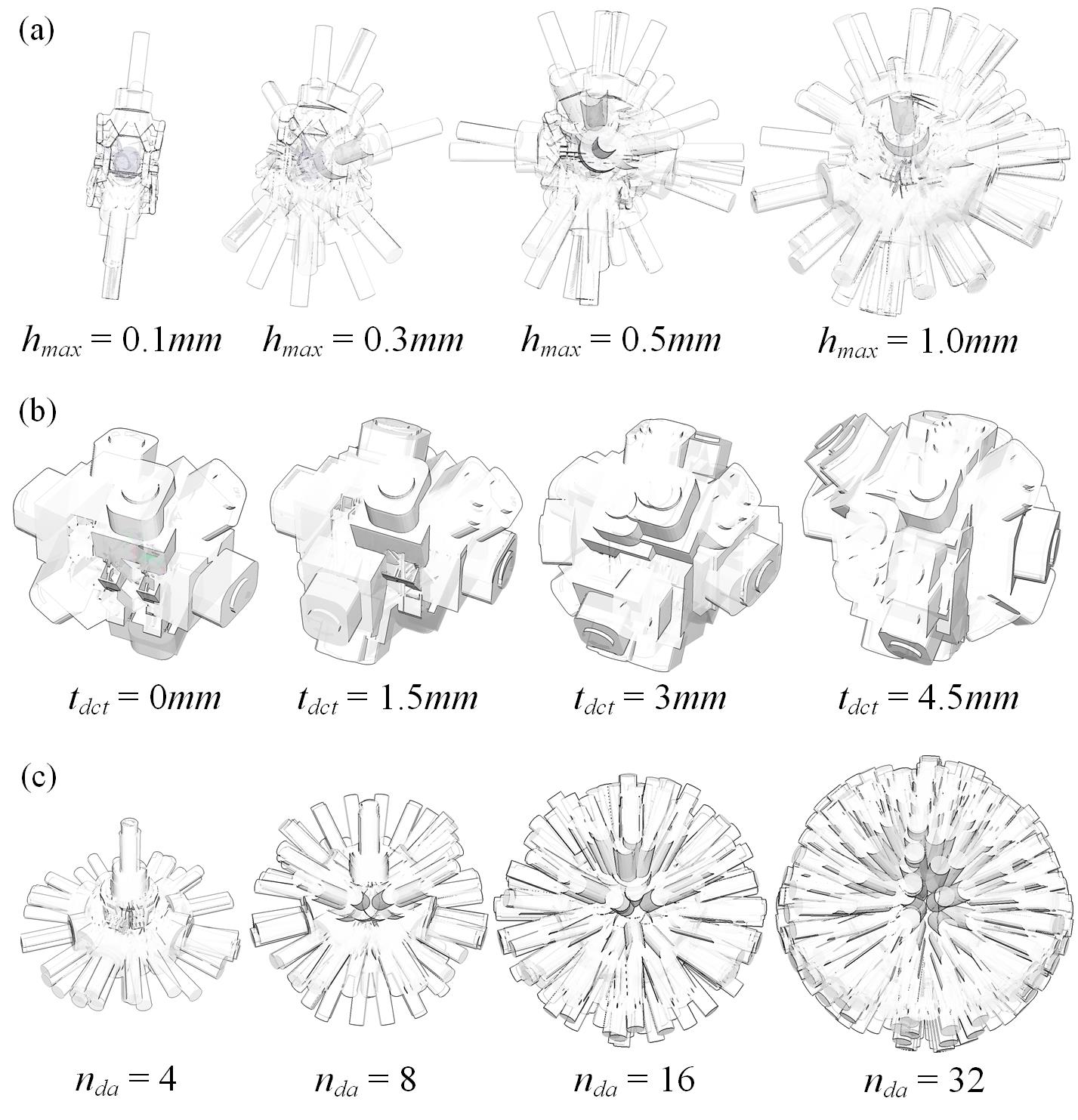}
  \caption{(a) Results of grasp planning using different $h_{max}$.
  As $h_{max}$ decreases, the hand grasps
  flat facets near the center of mass to maintain stability. 
  Object: Stanford bunny, the last model in Fig.\ref{costsgeneral}.
  (b) Results of different $t_{dct}$.
  As $t_{dct}$ increases, more grasps are found. Object: Tube connector, the fourth model in Fig.\ref{costsgeneral}. 
  (c) Results of different $n_{da}$.
  A larger $n_{da}$ leads to denser results. Object: Tube connector.}
  \label{tdctresults}
\end{figure}

\begin{figure*}[!htbp]
  \centering
  \includegraphics[width = \linewidth]{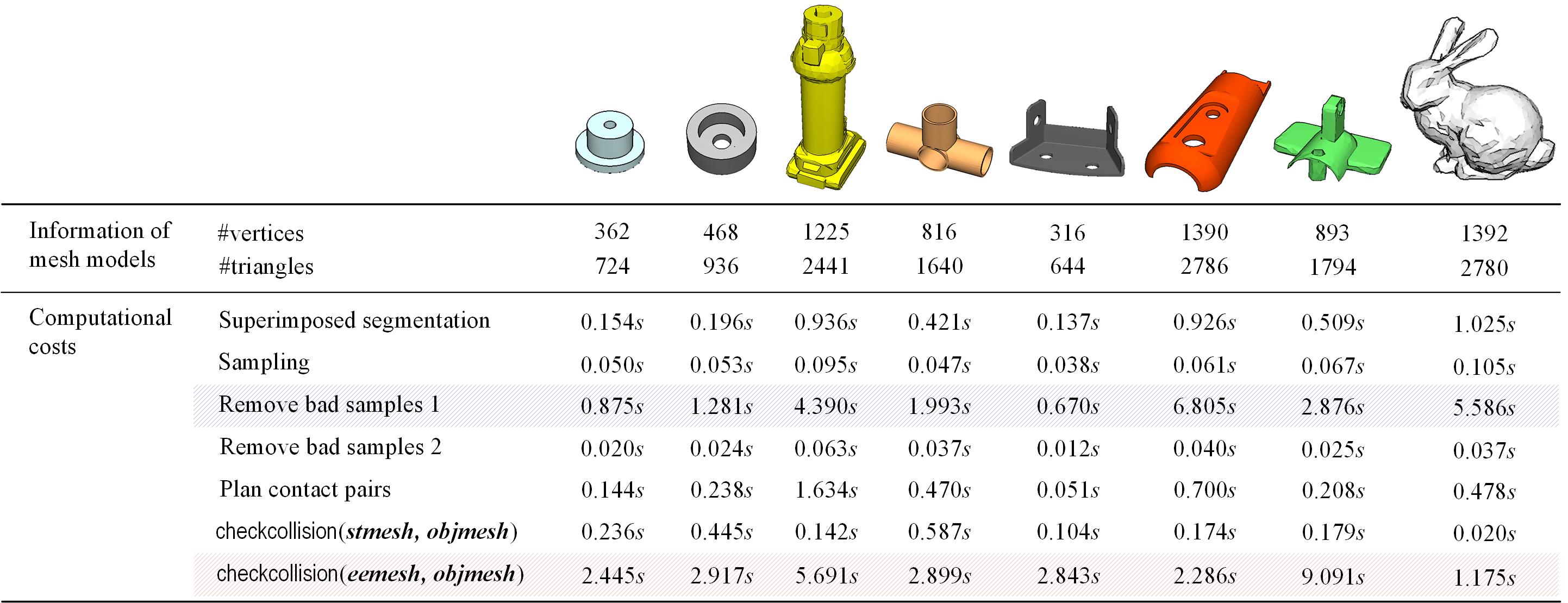}
  \caption{Computational cost of the proposed algorithms. The
  rows marked by red and blue shadows denote the most time consuming process. The
  values are the average results of ten executions using the following parameter
  setting: $\theta_{pln}$=$20^\circ$, $\theta_{fct}$=$20^\circ$,
  $t_{bdry}$=2$mm$, $t_{rnn}$=3$mm$, $t_{rss}$=50$mm$,
  $\theta_{parl}$=$160^\circ$, $t_{dct}$=3$mm$, $n_{da}$=8.}
  \label{costsgeneral}
\end{figure*}

\textbf{Parameter 1:} $\theta_{pln}$ is used to control the planarity of each
facet during the superimposed segmentation. Smaller $\theta_{pln}$ leads to flatter and smaller facets.
Fig.\ref{edsegresults}(a)
shows the segmented results of the electric drill shown in Fig.\ref{graspsdemo}(a) using different
$\theta_{pln}$. The facets are drawn with random offsets from
their original position to give a clear view. Each facet is given a random
color. As $\theta_{pln}$ becomes smaller, facets become
flatter and smaller. In the extreme case where $\theta_{pln}$ = $0^\circ$, each
triangle is treated as a facet.

\textbf{Parameter 2:} $\theta_{fct}$ is used to control the overlap of
superimposed segmentation. Smaller $\theta_{fct}$ leads to more overlap between
facets. Fig.\ref{edsegresults}(b) shows the segmented results of the tube
connector shown in Fig.\ref{threegraspsdemo}(a) using different $\theta_{fct}$.
Like Fig.\ref{edsegresults}(a), each facet is drawn with a random offset and a
random color. As $\theta_{fct}$ becomes smaller, facets become more overlapped.

\textbf{Parameter 3:} $t_{bdry}$ is used to control the distance between
contacts and facet boundaries. Fig.\ref{edsegresults1}(a) shows the results of
contact sampling using the metal workpiece shown in Fig.\ref{suctioncupdemo}(a)
and different $t_{bdry}$. Only samples on the bottom of the object are drawn and
the resulting samples are drawn in blue color.
A smaller $t_{bdry}$ leads to a smaller clearance between
contact samples and facet boundaries (hence less robust results).

\textbf{Parameter 4:} 
$t_{rnn}$ is used to control the radius of contact regions or the density of
contact sampling. Fig.\ref{edsegresults1}(b) shows the sampling results of the
metal workpiece using different $t_{rnn}$. The samples are drawn in red.
A smaller $t_{rnn}$ leads to denser contact samples on the object surface (hence more
planned grasps).

\textbf{Parameter 5:}
$h_{max}$ is used to control the stability of the planned grasps. An example is shown in 
Fig.\ref{tdctresults}(a). The object is the Stanford bunny (the last model in Fig.\ref{costsgeneral}).
As $h_{max}$ decreases, the planner reduces to grasp flat facets near the center of mass
to maintain stability.

\textbf{Parameter 6:} 
$\theta_{parl}$ is used to control the parallelity of the facet pairs in the grasp
planning. Values with a larger offset from $180^\cdot$ lead to more candidate ``parallel''
facet pairs to attach the finger pads and hence more planned grasps. On the other
hand, values with larger offsets result in unstable grasping configurations.
Fig.\ref{edsegresults2} shows the parallel facets of a toy wheel using different
$\theta_{parl}$. Arrows indicate facet normals. Two facets with
the same arrow color are parallel. Like Fig.\ref{edsegresults}(a), the facets are
drawn with random offsets from their original position to give a clear view. As
$\theta_{parl}$ becomes smaller, ``parallel'' facets become less parallel.
Meanwhile, the number of parallel facets becomes larger.

\textbf{Parameter 7:} 
$t_{dct}$ is used to control the distances between finger pads and object
surfaces in the grasp planning for three-finger grippers.
Larger values indicate the planner allows a large difference in
distances between finger pads and object surfaces. In that case, there will be
more planned grasps. Meanwhile, the results are less robust since two fingers
cannot touch object surfaces at the same time.
Fig.\ref{tdctresults}(b) shows the planned grasps of the tube
connector shown in Fig.\ref{threegraspsdemo}(b) using different
$t_{dct}$. As $t_{dct}$ becomes larger, the planned grasp configurations become
denser. The object, which is obstructed by hands in the figure, is at the same
pose as Fig.\ref{threegraspsdemo}(b). The results also show that when $t_{dct}$
equals 0$mm$, there are no lateral grasps. As $t_{dct}$ becomes larger, the
number of lateral grasps increase. 

\textbf{Parameter 8:}
In addition, $n_{da}$ determines the number
of discretized rotation angles around contact normals. A larger  $n_{da}$
leads to denser results, as is shown in Fig.\ref{tdctresults}(c).

\subsection{Performance}

\subsubsection{Computational costs}

The computational costs of planning grasps for various objects using the method
are shown in Fig.\ref{costsgeneral}. Six objects are used. From left to right, 
they are (1) a bearing housing, (2) a toy
wheel, (3) an electric drill, (4) a tube connector, (5) a metal workpiece, and
(6) a plastic workpiece. The information of these mesh models, including the
number of vertices and the number of triangle faces, is shown in an upper
section of the table in Fig.\ref{costsgeneral}. The details of computational cost,
including the time spent on superimposed segmentation, sampling, removing bad
samples 1 (refinement 1 of Fig.\ref{refinements}), removing bad samples 2
(refinement 2 of Fig.\ref{refinements}), planning contact pairs, and the two
nested collision detection, are shown in a lower section of the table in
Fig.\ref{costsgeneral}. The results are obtained by running the algorithms on a
LENOVO ThinkPad P70 mobile workstation. One core of an Intel
Xeon E3-1505M v5 @ 2.80GHz 4 Core CPU is used. The memory size is 16.0GB. The
algorithms are implemented using python 2.7.11 32 bit. The
results are the average values of ten executions using the following parameter
settings: $\theta_{pln}$=$20^\circ$, $\theta_{fct}$=$20^\circ$,
$t_{bdry}$=2$mm$, $t_{rnn}$=3$mm$, $h_{max}$=1.5$mm$,
$\theta_{parl}$=$160^\circ$, $t_{dct}$=3$mm$, $n_{da}$=8.

The top two time-consuming rows are marked by red and blue shadows in
Fig.\ref{refinements}. The first one is
$\mathtt{checkcollision}(\textit{\textbf{eemesh}}, \textit{\textbf{objmesh}})$,
which required a few seconds for a few thousand triangles. The results are
reasonable as we are performing mesh-to-mesh collision detections. The
second one is ``Remove Bad Samples 1'' (see Section III.B.2) and Fig.\ref{refinements}).
The cost is also reasonable since it measures the distances of each sampled 
contact point to the boundaries of facets. 

\subsubsection{Influence of mesh quality}

\begin{figure*}[!htbp]
  \centering
  \includegraphics[width = \linewidth]{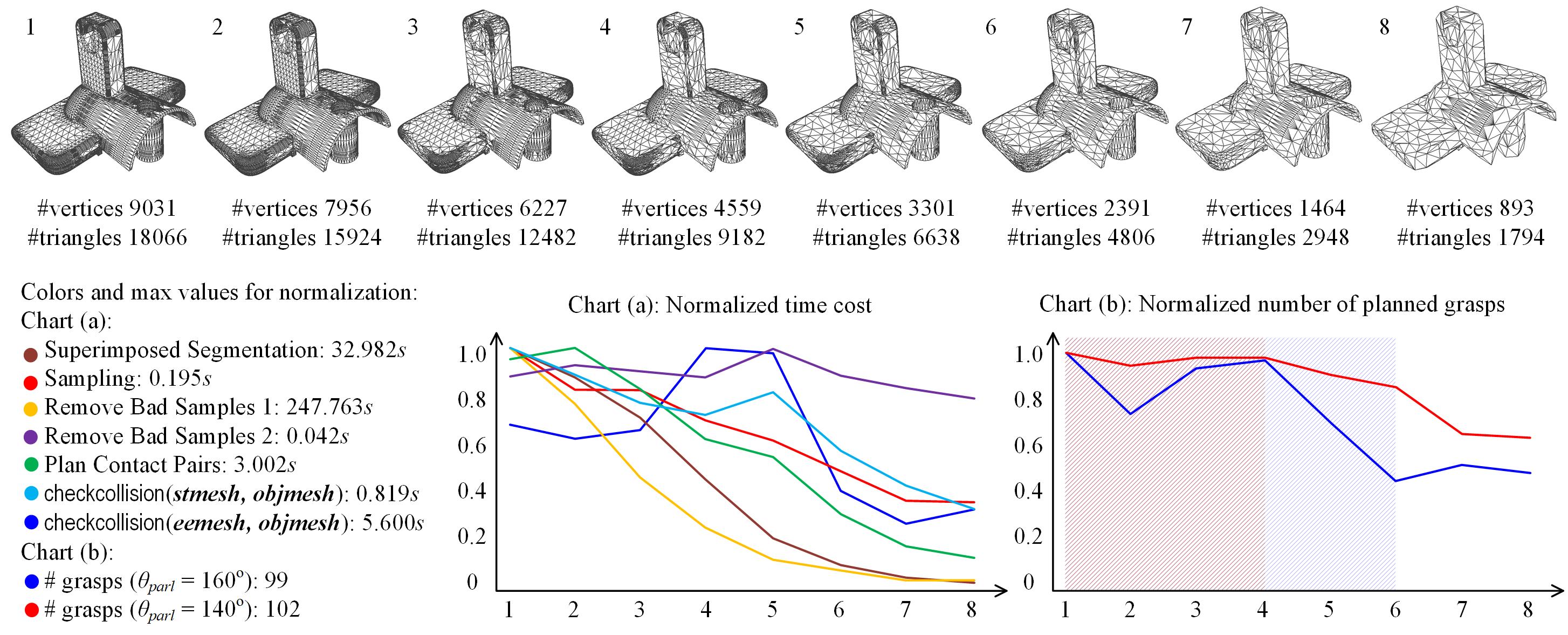}
  \caption{Performance of the proposed algorithms using a model with decreasing
  mesh qualities. The meshes in 1$\sim$8 have a decreasing number
  of vertices and triangles and thus decreasing qualities. The curves in the
  two charts show the changes in time costs and the number of planned grasps as
  the mesh quality decreases.}
  \label{costchanging}
\end{figure*}

Fig.\ref{costchanging} shows the performance of the algorithms using a model with different
mesh qualities. The meshes are drawn in	
Fig.\ref{costchanging}(1$\sim$8). The number of
vertices and triangles of the meshes decreases from 1 to 8. The two charts in the
figure show the normalized time cost and the normalized number of planned grasps on these models.
Here, the maximum values
are normalized as 1, the other values are normalized to values between 0 and 1.
The maximum values are used as denominators for normalization. They are listed in
the lower-left corner of the figure. In Chart (a), the different curves
show the changes of various cost. The meanings of the colors are also shown in
the lower-left corner. All curves in (a) are the results of the
same parameter settings: $\theta_{pln}$=$20^\circ$, $\theta_{fct}$=$20^\circ$,
$t_{bdry}$=2$mm$, $t_{rnn}$=3$mm$, $h_{max}$=1.5$mm$,
$\theta_{parl}$=$160^\circ$, $t_{dct}$=3$mm$, $n_{da}$=8. The two curves in Chart
(b) show the number of planned grasps. The red curve is the changes of grasp number using
the same parameter setting. The blue curve shows the
results using a different $\theta_{parl}$ value ($\theta_{parl}$=$140^\circ$). The
two curves show that the algorithms are stable to low-quality mesh models: The
number of planned grasps does not significantly decrease along with reduced
vertices and triangles. For $\theta_{parl}$=$160^\circ$, the number of
planned grasps is considered to have similar values in the red
shadow (spans from 1 to 4). For $\theta_{parl}$=$140^\circ$, the number of
planned grasps is considered to have similar values in the blue
shadow (spans from 1 to 6).

\subsubsection{Precision of the planned grasps}
\begin{figure}[!htbp]
  \centering
  \includegraphics[width = \linewidth]{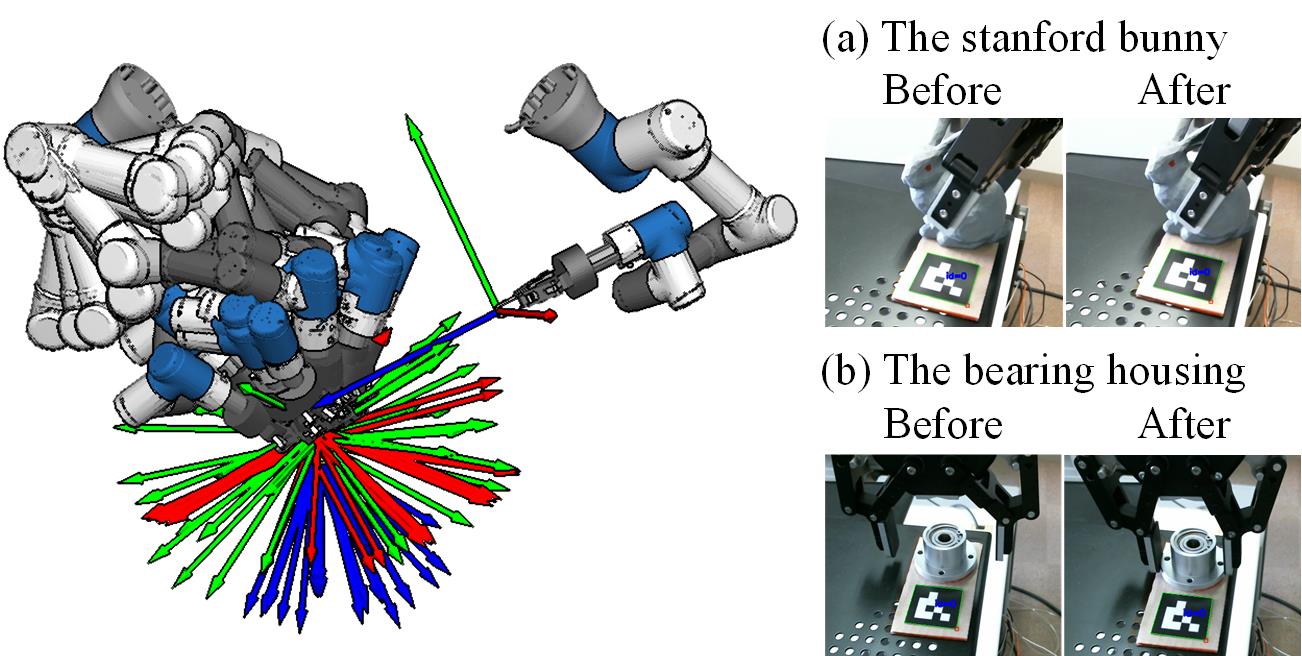}
  \caption{Experimental settings used to examine the precision of the planned
  grasps. The left part shows the grasps with approaching directions that have
  less than $40^\circ$ angle from the vertical direction. The changes in object
  poses before and after grasping are measured by AR markers shown in the right.
  Two objects, a Stanford bunny and a bearing housing, are tested.}
  \label{preicisionsetup}
\end{figure}

We further measure the precision of the planned grasps using a
Robotiq85 two-finger parallel gripper and some objects.
The experiment settings are shown in the right part of
Fig.\ref{preicisionsetup}. The objects include (a) the Stanford bunny, and (b)
the bearing housing. AR markers are attached to the objects to
precisely detect the changes of poses before and after closing the
fingers. The grasps with approaching directions that have less than 40 degree
angle from the vertical direction, as shown in the left part of
Fig.\ref{preicisionsetup}, are selected as the candidate grasps.
They are used to grasp the objects. The difference in object poses
(the difference in the AR marker's $x$ and $y$ positions, namely $d_x$ and $d_y$)
before and after closing the fingers are measured as the precision.

\begin{figure}[!htbp]
  \centering
  \includegraphics[width = \linewidth]{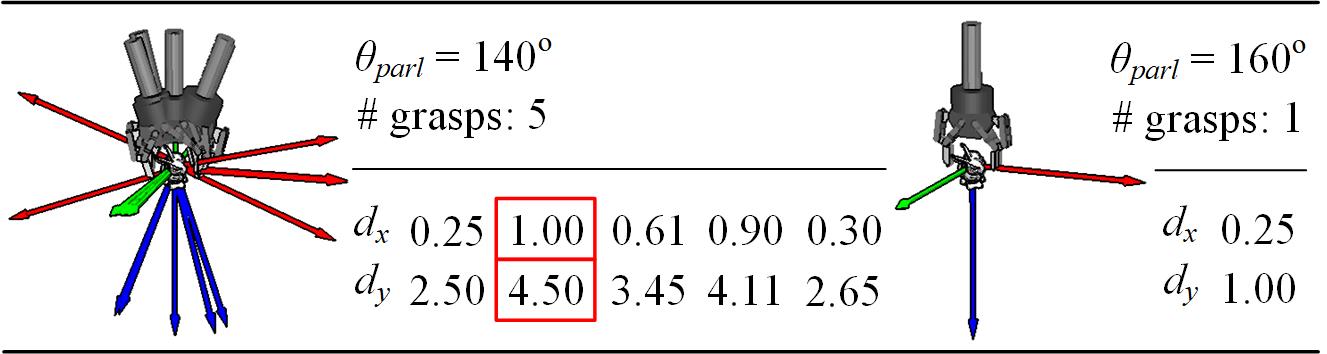}
  \caption{Changes in positions after grasping the Stanford bunny using
  the planned grasps. The small 3D figures show the candidate grasps used for
  comparison. When $\theta_{parl}$=$140^\circ$, 6 candidate grasps are found. They
  are shown in the left 3D figure. When $\theta_{parl}$=$160^\circ$, only 1
  candidate grasp is found. It is shown in the right 3D figure.}
  \label{bunny}
\end{figure}

The results of the Stanford bunny after grasping are shown
in Fig.\ref{bunny}. In the left case, $\theta_{parl}$ was set
to $140^\circ$, and 6 candidate grasps were found. The maximum change after
grasping using these planned grasps was $d_x$=1.00$mm$ and $d_y$=4.5$mm$.
When $\theta_{parl}$ was changed to $160^\circ$ (the right part of Fig.\ref{bunny}), 
only 1 candidate grasp was found.
Its change was $d_x$=0.25$mm$ and $d_y$=1.00$mm$.

\begin{figure*}[!htbp]
  \centering
  \includegraphics[width = \linewidth]{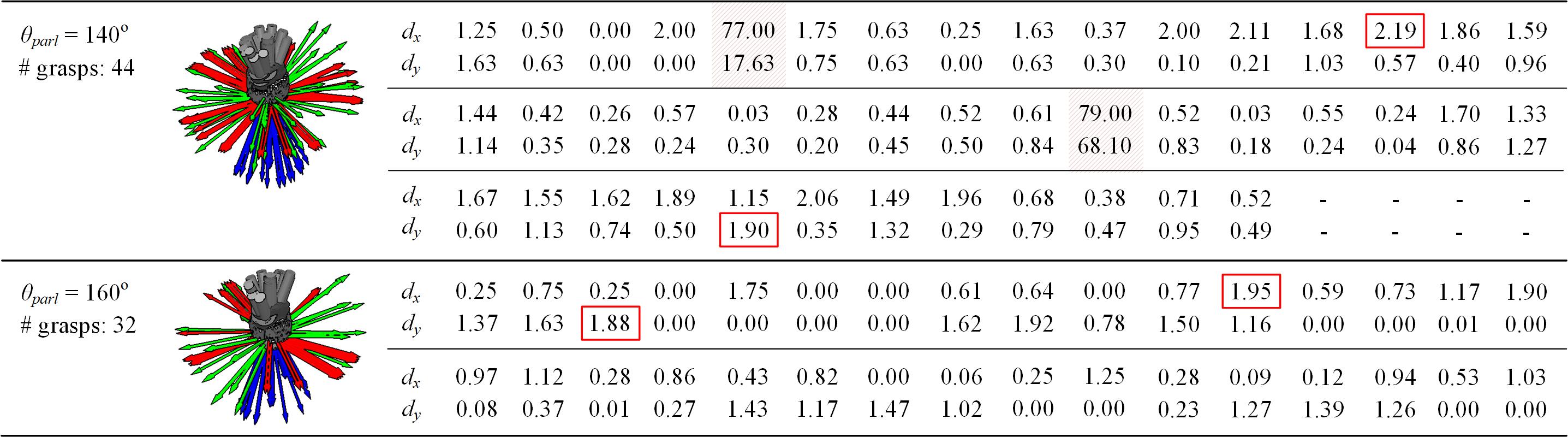}
  \caption{Changes in $x$ and $y$ after grasping the bearing housing using
  planned grasps. When $\theta_{parl}$=$140^\circ$, 44 candidate grasps were found.
  Their poses and precision are shown in the upper section of the table. The maximum
  positions change after grasping were $d_x$=2.19$mm$ and $d_y$=1.90$mm$. They
  are marked using red frames. Two failures were encountered when grasping using
  these planned grasps. They are marked using red shadows. When 
  $\theta_{parl}$=$160^\circ$, 32 candidate grasps were found. Their poses and
  precision are shown in the lower section. The maximum changes were
  $d_x$=1.95$mm$ and $d_y$=1.88$mm$.}
  \label{preicision}
\end{figure*}

\begin{figure*}[!htbp]
  \centering
  \includegraphics[width = \linewidth]{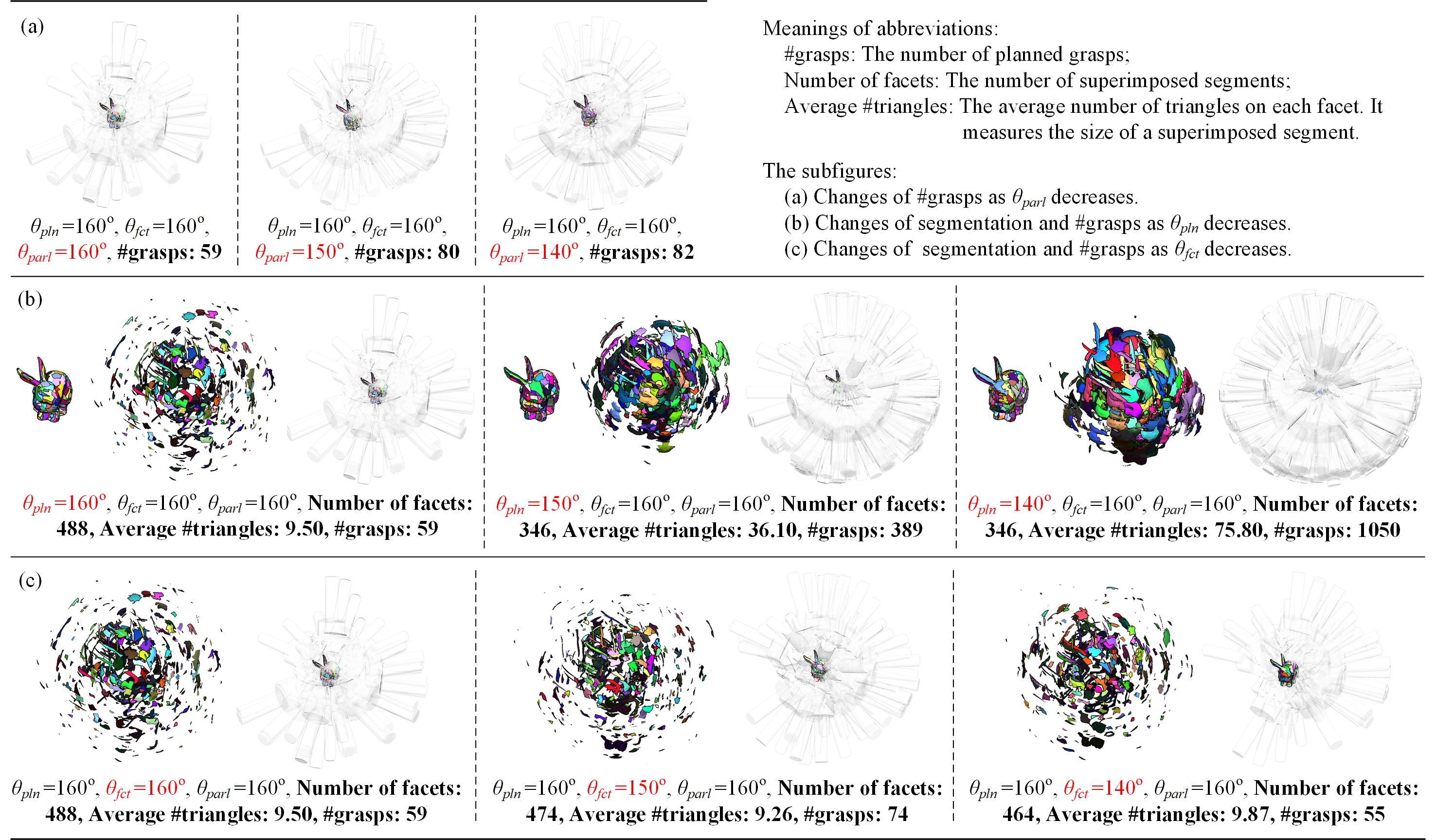}
  \caption{Changes of segmentation and planned grasps of the stanford
  bunny with different parameter settings. The parameters included
  $\theta_{pln}$, $\theta_{fct}$, and $\theta_{parl}$. At each row, two of these
  parameters are fixed to $160^\circ$, the left one decreases from left to right.
  }
  \label{bunnydetails}
\end{figure*}

The Stanford bunny does not have many graspable parallel pairs. The available
data might not be enough to demonstrate the precision of the planned grasps.
Thus, we further examined the planned grasps of the bearing housing.
The $d_x$ and $d_y$ before and after grasping the bearing housing using the
planned grasps are shown in Fig.\ref{preicision}.
The first two rows of the table are the results using
parameter settings: $\theta_{pln}$=$20^\circ$, $\theta_{fct}$=$20^\circ$,
$t_{bdry}$=2$mm$, $t_{rnn}$=3$mm$, $h_{max}$=1.5$mm$,
$\theta_{parl}$=$140^\circ$, $t_{dct}$=3$mm$, $n_{da}$=8.
The lower two rows are the results using a different $\theta_{parl}$. $\theta_{parl}$=$160^\circ$. When
$\theta_{parl}$=$140^\circ$, there are 44 candidate grasps. Two failures are
encountered during the experiments using these planned grasps.
The failure cases are marked in red shadows. When $\theta_{parl}$=$160^\circ$,
there are 32 candidate grasps, and all of them can successfully hold the
object. The maximum position change after grasping are ($d_x$=2.19$mm$,
$d_y$=1.90$mm$), and ($d_x$=1.95$mm$, $d_y$=1.88$mm$) in the two cases
respectively. They are marked in red frames. With the results of these two
models, we confirm that the planned grasps have satisfying precision and are
suitable to be used by assembly routines (e.g. spiral search). Also, we may change the
parameters of the algorithms and seek a balance between precision and the number
of planned grasps according to the requirements of specific tasks (industrial bin-picking
or assembly).

For readers' convenience, we summarize the detailed changes of segmentation and
planned grasps of the Stanford bunny with different parameter settings in
Fig.\ref{bunnydetails}. In (a), $\theta_{pln}$ and $\theta_{fct}$ are fixed to
$160^\circ$, $\theta_{parl}$ decreases from left to right. The number of planned
grasps increases as $\theta_{parl}$ decreases. In (b), $\theta_{fct}$ and
$\theta_{parl}$ are fixed to $160^\circ$, $\theta_{pln}$ decreases from left to
right. The sizes of facets increase significantly as $\theta_{pln}$ decreases
(The sizes of facets are measured by the average number of triangles).
Meanwhile, as the sizes of facets increases, the number of planned grasps
increases significantly. In (c), $\theta_{pln}$ and $\theta_{parl}$ are fixed
to $160^\circ$, $\theta_{fct}$ decreases from left to right. The number of facets
decreases slightly as $\theta_{fct}$ decreases. There are no necessary relations
between $\theta_{fct}$ and the size of facets, and between $\theta_{fct}$ and the
number of planned grasps.


\begin{figure*}[!htbp]
  \centering
  \includegraphics[width = \linewidth]{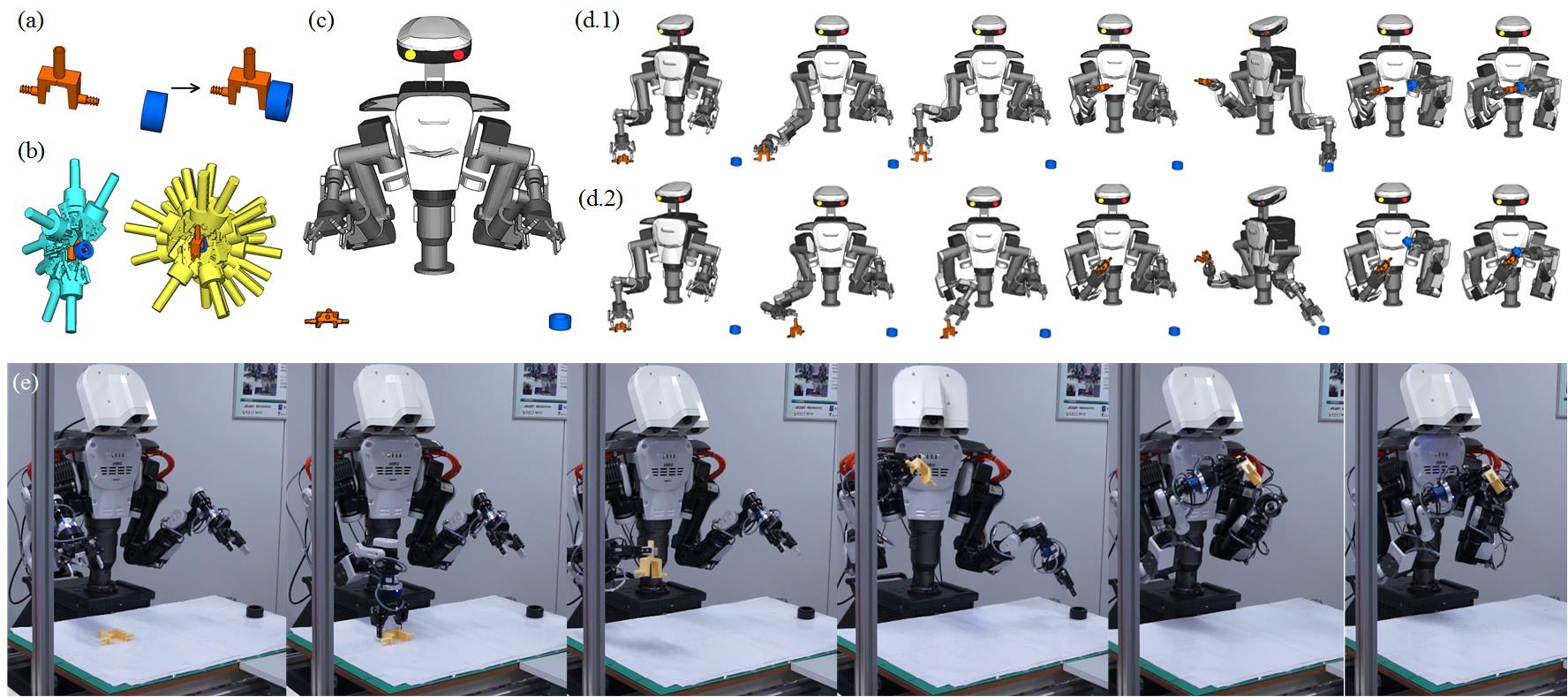}
  \caption{Using a Kawada Nextage robot
  to conduct an assembly task.
  (a) The goal of the assembly task. (b) The planned
  collision-free grasps for assembly. The
  cyan hands show the possible grasp configurations to hold the support. The
  yellow hands show the possible grasp configurations to hold the wheel. (c) The
  initial poses of the objects to be assembled. (d) Two exemplary planned
  sequences using the planned grasps. (e) A real-world execution.}
  \label{regrasp}
\end{figure*}

\subsection{Some robotic assembly examples using the grasp planner}

The proposed method is implemented as a plugin of an open source project named
PYHIRO (available on Github at https://github.com/wanweiwei07/pyhiro), where the
goal is to build a system that conducts assembly tasks without
human teach. The input to the system is the CAD models of objects and kinematic
parameters of robots and hands. The output is a sequence of grasp configurations
and robot poses for assembly. The proposed grasp planning is a pre-processing
component of the project. It prepares pre-annotated grasp configurations for
regrasp planning and motion planning. 
This subsection presents some examples of robotic
assembly using the systems and the proposed grasp planning algorithms.


The first example is to use a Kawada Nextage robot (Kawada Industries, Inc.) to assemble a wheel 
to a support shown in Fig.\ref{regrasp}(a). The hands used were two Robotiq F-85 grippers. The
input is the initial positions and orientations of the wheel and the support in
a robot's workspace (the initial poses are shown in Fig.\ref{regrasp}(c)). The
system automatically plans grasps, invalidates collided grasp configurations,
and plans motion sequences using the collision-free grasps.
The planned collision-free grasps for assembling the two objects are
shown in Fig.\ref{regrasp}(b). Two exemplary planned sequences are shown in
Fig.\ref{regrasp}(d.1) and (d.2). The real-world execution using one sequence is
shown in Fig.\ref{regrasp}.
The planned grasps are precise enough to guarantee successful pick-and-place and
regrasp.

\begin{figure}[!htbp]
  \centering
  \includegraphics[width = \linewidth]{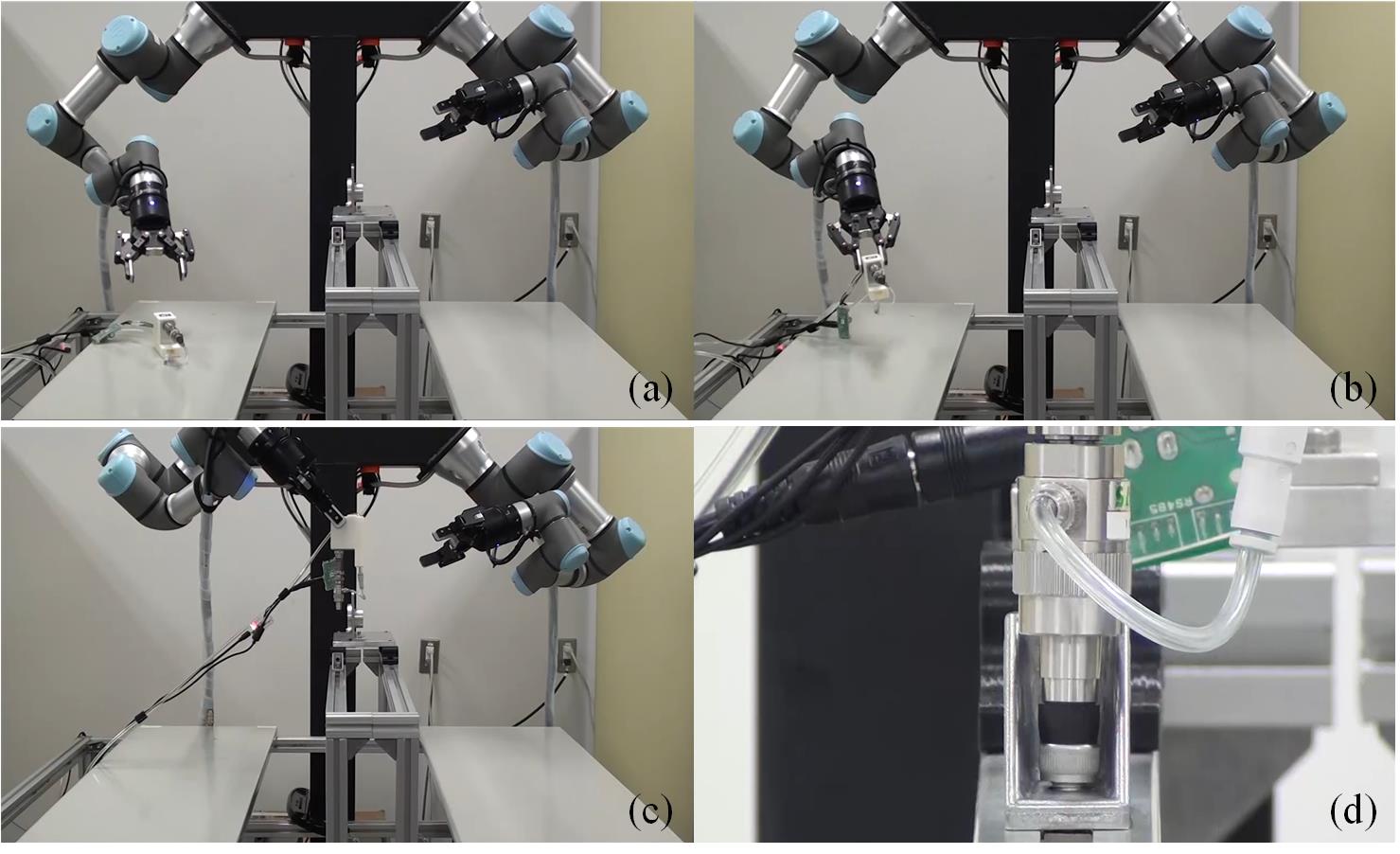}
  \caption{Using a dual-arm UR3 robot to pick up a vacuum fastener 
  and fasten a bolt. (a-c) The robot chooses a suitable grasp from preplanned grasps
  to pick up the vacuum fastener.
  (d) The planned grasp is precise enough to align the
  suction tooltip and fasten the bolt.}
  \label{driver}
\end{figure}

The second example is to use a dual-arm UR3 robot (Universal Robots A/S) to pick up a vacuum fastener 
and fasten a bolt. The robot plans a bunch of candidate grasps using the 
proposed algorithms, and chooses a suitable grasp to pick up the vacuum fastener
in Fig.\ref{driver}(a-c). The planned grasp is precise
enough to align the suction tooltip and fasten the bolt in Fig.\ref{driver}(d).
Like the first example, the gripper used is the
Robotiq F-85 gripper.

\begin{figure}[!htbp]
  \centering
  \includegraphics[width = \linewidth]{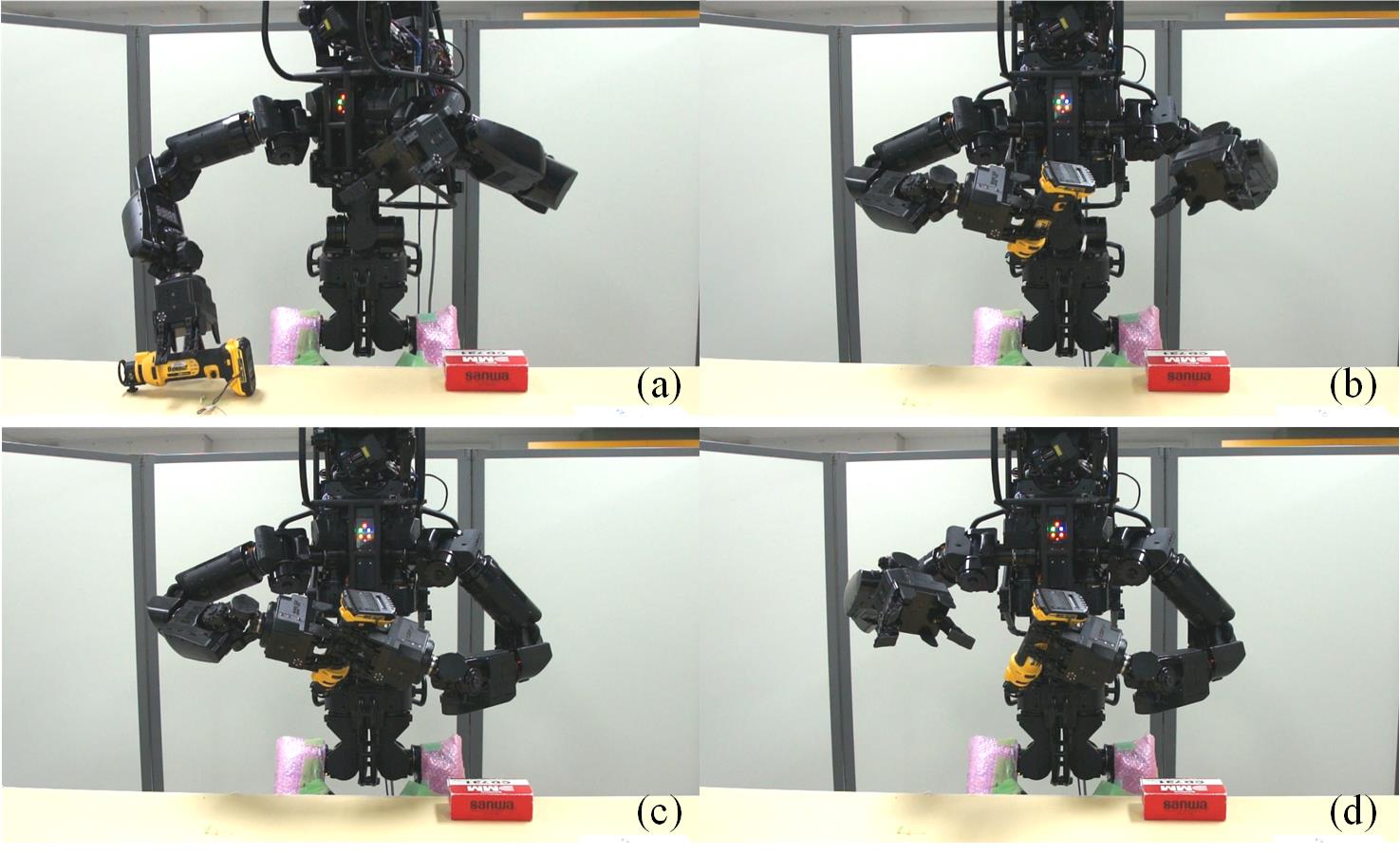}
  \caption{Handing over a tool driver from the right hand of an HRP5P
  robot to its left hand. (a-b) The robot picks up the electric drill using its left hand
  and prepares it for handover. (c) Handover. The right hand inserted its thumb into
  the narrow space between the index and middle fingers of the right hand to hold the drill.
  (d) The left hand released and the object is successfully delivered.}
  \label{handover}
\end{figure}

The third example is to hand over an electric drill from the right hand of an HRP5P
robot (BNational Inst. of AIST, Japan, http://y2u.be/ARpd5J5gDMk) to its left hand. The gripper is a three-finger parallel gripper. The
robot could precisely hold the object using the planned grasps (Fig.\ref{handover}(a-b)), 
which allows the left hand to insert its thumb into
the narrow space between the index and middle fingers of the right hand Fig.\ref{handover}(c-d).

\section{Conclusions and Future Work}

In this paper, we proposed grasp synthesis algorithms that are efficient, have
high precision, and are highly configurable with several tunable parameters. The
algorithms focused on industrial end-effectors like grippers and suction cups,
and can be used in industrial tasks like bin-picking and assembly. The
efficiency, precision, and configurability of the proposed method
meet the requirements of these tasks.
The proposed planner is demonstrated to be practical by a real-world robotic assembly task.

We envision our future research focusing on non-continuous contacts. In this paper,
a finger pad is assumed to be continuously in contact with a flat surface of an
object, which makes it difficult to find grasps on non-continuous surfaces,
e.g. grasp the screw threads of a bolt. In the future, we will use surface
simplification, reconstruction, and multi-contact analysis to plan grasps and
challenge the difficulty.

\bibliographystyle{ieeetr}
\bibliography{reference}

\end{document}